\documentclass[journal]{IEEEtran}

\usepackage{graphicx}
\usepackage{cite}
\usepackage{amssymb}
\usepackage{latexsym,amscd}

\usepackage{graphicx}
\usepackage{float}

\usepackage{times}

\usepackage{soul}
\usepackage{url}
\usepackage[hidelinks]{hyperref}
\usepackage[utf8]{inputenc}
\usepackage[small]{caption}
\usepackage{graphicx}
\usepackage{amsmath}
\usepackage{booktabs}
\usepackage{threeparttable}  
\usepackage{xcolor}

\usepackage{subfigure} 
\usepackage{multirow}
\usepackage{algorithm}
\usepackage{algorithmic}

\usepackage{ifthen}
\usepackage{float}
\usepackage{amssymb}

\usepackage{color}

\ifCLASSINFOpdf
\else
\fi

\begin{document}

\title{Detail Reinforcement Diffusion Model: Augmentation Fine-Grained Visual Categorization in Few-Shot Conditions}

\author{Tianxu~Wu, Shuo~Ye, Shuhuang~Chen, Qinmu~Peng and Xinge~You,~\IEEEmembership{Senior Member,~IEEE}

\thanks{This work was supported in part by the National Key R\&D Program of China 2022YFC3301000, in part by the Fundamental Research Funds for the Central Universities, HUST: 2023JYCXJJ031. Co-corresponding author: \textit{ Shuo~Ye(shuoye.ke@gmail.com), Qinmu Peng(e-mail: pqinmu@gmail.com)} }

\thanks{TianXu~Wu, Shuo~Ye, and Shuhuang Chen are with the School of Electronic Information and Communications, Huazhong University of Science and Technology, Wuhan 430074, China.}

\thanks{Qinmu~Peng and Xinge~You are with the School of Electronic Information and Communications, Huazhong University of Science and Technology, Wuhan 430074, China.}

\thanks{©2024 IEEE. Personal use of this material is permitted.   Permission from IEEE must be obtained for all other uses, in any current or future media, including reprinting/republishing this material for advertising or promotional purposes, creating new collective works, for resale or redistribution to servers or lists, or reuse of any copyrighted component of this work in other works.}
}

\maketitle

\begin{abstract}
The challenge in fine-grained visual categorization~(FGVC) lies in how to explore the subtle differences between different subclasses and achieve accurate discrimination.
Previous research has relied on large-scale annotated data and pre-trained deep models to achieve the objective. However, when only a limited amount of samples is available, similar methods often struggle to accurately learn the details of instances and perform effective recognition.
Using diffusion models for data augmentation has gained widespread attention, but the high level of detail required for fine-grained images makes it challenging for existing methods to be directly employed.
To address this issue, we propose a novel approach termed the detail reinforcement diffusion model~(DRDM), which leverages the extensive knowledge of large models for fine-grained data augmentation and comprises two key components including discriminative semantic recombination (DSR) and spatial knowledge reference~(SKR).
Specifically, DSR is designed to extract implicit similarity relationships from the labels and reconstruct the semantic mapping between labels and instances, which enables better discrimination of subtle differences between different subclasses.
Furthermore, we introduce the SKR module, which incorporates the distributions of different datasets as references in the feature space. This allows the SKR to aggregate the high-dimensional distribution of subclass features in few-shot FGVC tasks, thus expanding the decision boundary.
Through these two critical components, we effectively utilize the knowledge from large models to address the issue of data scarcity, resulting in improved performance for fine-grained visual recognition tasks.  
Extensive experiments demonstrate the consistent performance gain offered by our DRDM.
\end{abstract}


\IEEEpeerreviewmaketitle

\section{Introduction}

\IEEEPARstart
{F}{i}ne-grained visual categorization (FGVC) aims to achieve the recognition of subclasses that exhibit tiny visual distinctions within the same large class (e.g., birds~\cite{wah2011caltech}). Related studies have been extensively applied to autonomous vehicles~\cite{8591961,sadeghi2020system} and pharmaceutical products~\cite{yi2022pharmaceutical,9870199}. Compared to general images, fine-grained images usually have similar features and are affected by interferences such as posture, perspective, and occlusion~\cite{ye2023image}. Therefore, the key to achieving FGVC often lies in discovering discriminative regions. This process is often achieved through the localization branch network~\cite{he2018stackdrl,zheng2020fine} or implicitly learned in end-to-end training~\cite{ding2019selective,ye2022discriminative,wang2022r2}. While automatically identifying these regions from a large-scale labeled dataset is feasible, many practical FGVC tasks lack such datasets because annotating fine-grained data is time-consuming, and labeling rare subclasses demands experienced expertise.  For instance, in the medical domain, discerning subtle feature differences among different subtypes of diseases, or in the industrial sector, identifying minute variations among components, and in ecology, recognizing specific types of pests or diseases are particularly reliant on expert annotation.
The ability of deep neural networks to process fine-grained few-shot learning~(FSL) is crucial for practical applications. Unfortunately, existing methods still perform much worse than weakly supervised methods on several few-shot benchmarks~\cite{hong2022semantic}. Networks often struggle to select the correct regions for recognition and tend to overfit pseudo-features from the training data~\cite{shu2022improving}.

\begin{figure}[t]
  \begin{center}
  \includegraphics[width=0.49\textwidth]{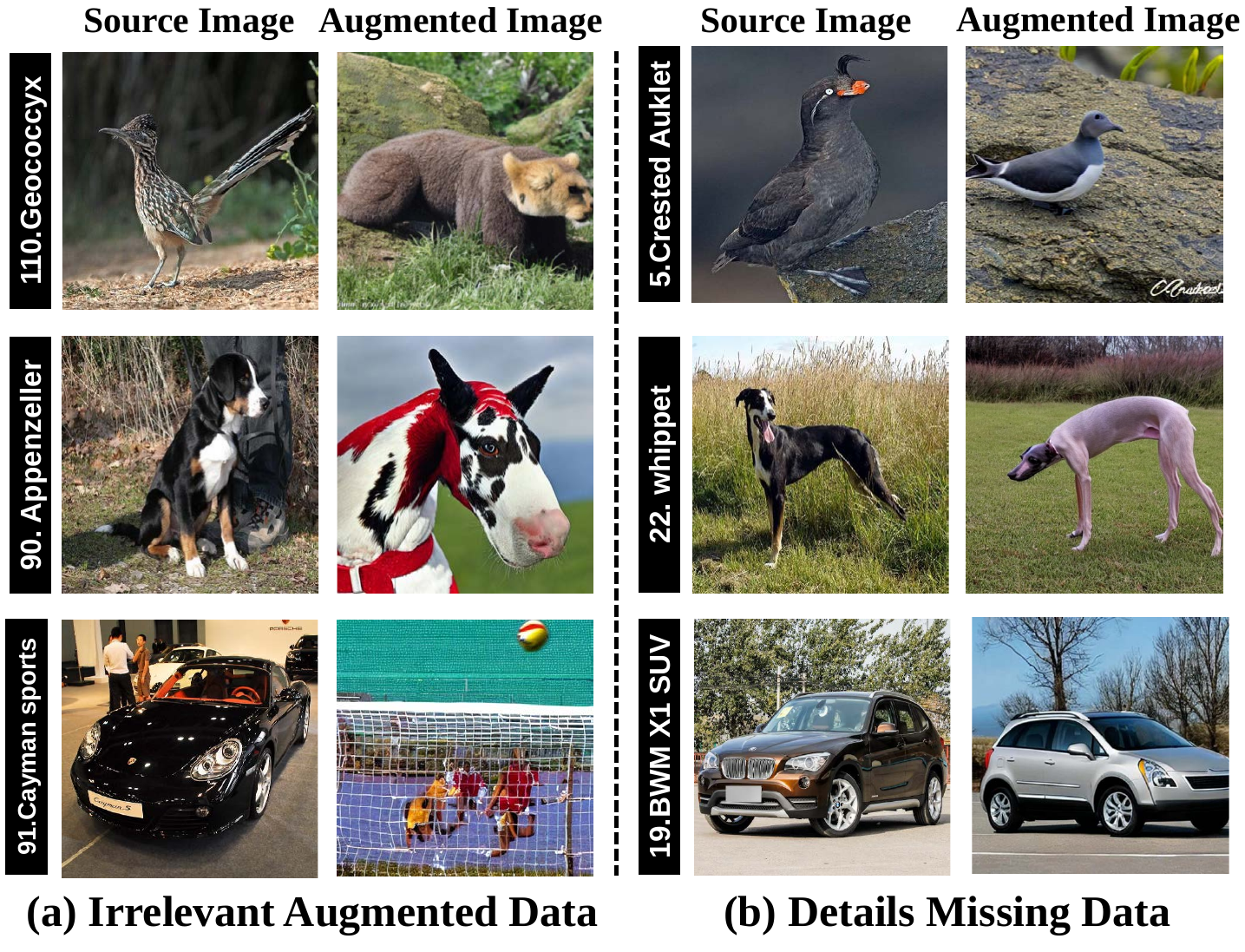}
  \end{center}
	\setlength{\abovecaptionskip}{-0.1cm} 
   \caption{Feature contamination resulting from semantic misalignment during data augmentation using large models. This is specifically evident in the form of (a) irrelevant augmented data and (b) the loss of discriminative details.}
\vspace{-0.5cm} 
  \label{motivation}
\end{figure}

Utilizing external information~(e.g., multi-view\cite{10234694} or multi-party\cite{10115455} information) can significantly enhance the performance of FSL, but this involves complex information acquisition pathways. A direct method to mitigate overfitting in FSL is through data augmentation~\cite{li2020adversarial}.
However, reliably getting diverse data remains a challenging problem, where the augmented instances should contain discriminative features of the classes and exhibit high intra-class diversity~\cite{le2019shadow}. Unfortunately, this often breaks down in adversarial learning~\cite{zhang2018metagan,gao2018low,tsutsui2019meta} methods, where there is a shortfall in generating diverse samples.
Recently, leveraging prior knowledge from large models (e.g., stable diffusion~\cite{rombach2022high}) for data augmentation demonstrates significant potential.
However, this success has not seamlessly extended into the realm of FGVC, one reason to consider is feature contamination, as depicted in Figure~\ref{motivation}.
It is manifested specifically as augmented data being unrelated to the original data or suffering from detail loss. In (a), when the label \textit{Geococcyx} is used as input, the model fails to generate the expected result, the result is an unrelated composite animal image. Similarly, in (b), when the input image is \textit{Crested Auklet}, although the generated images possess bird-like structures, their detailed features do not align with the target subclass.
Please note that this phenomenon has been observed across different types of datasets.
One of the reasons contributing to this phenomenon is believed to be the specialized nature of fine-grained labels. 
This implies that nouns within these labels are less common compared to general images. 
Consequently, during the pre-training process, this inherent imbalance poses a challenge for models to effectively learn fine-grained information and accurately establish a mapping between labels and semantic features. As a result, they struggle to depict fine-grained features during data augmentation with large models.
Utilizing such feature contamination images for training fine-grained models would severely impair the models' understanding of instances. 
Moreover, limited instances can also result in data feature points struggling to encompass the intricate boundaries between different categories. Models may lean towards adopting simplistic decision boundaries, preventing them from capturing the complex classification scenarios present in the real world.

We argue that since fine-grained labels possess a certain level of expertise, the naming of subclasses should adhere to specific conventions.
For instance, labels like \textit{Parakeet Auklet} and \textit{Crested Auklet} both contain \textit{Auklet} in their names. This kind of textual similarity implicitly encodes fundamental subclass features, such as a red beak and a short tail.
By utilizing the inherent resemblance in label descriptions, the data augmentation process can be constrained, thereby effectively enhancing the performance of FGVC in few-shot conditions. To address this, we propose a detailed reinforcement model.
Specifically, the discriminative semantic recombination module is designed to explicitly emphasize subclass-specific differences from a labeling perspective. It then utilizes the extracted similarity relationships to guide and constrain the data augmentation process performed by diffusion models.
Meanwhile, the spatial knowledge reference module is designed to incorporate diverse data distributions from various data types as reference points into the feature space. This approach effectively addresses the challenge of poorly defined decision boundaries in FSL due to limited data, thereby enhancing the model's instance understanding.
Our model demonstrates notable scalability and can seamlessly integrate knowledge supplementation from different data modalities, leveraging reference knowledge from datasets of distinct types.
Our main contributions are summarized as follows:

\begin{itemize}
\item We analyzed the limitations of applying the diffusion model to fine-grained image data augmentation and proposed a Discriminative Semantic Recombination (DSR) module. This module effectively explores the relationships between instance labels and image information under weakly supervised conditions, thus enhancing the details of augmented data.

\item We proposed a Spatial Knowledge Referencing~(SKR) approach that introduces the distributions of different data types as references in the feature space. This encourages the model to find clear and distinct boundaries for fine-grained features in the high-dimensional space, thereby enhancing the model's understanding of instances.

\item We conducted extensive experiments on three benchmark datasets, and the results demonstrated that the proposed DRDM achieved favorable performance in the FSFG problem, and significantly improved the overall performance compared to other methods.
\end{itemize}


\section{Related Work}

\subsection{Fine-Grained Visual Categorization in Few-Shot Setting}
Fine-grained visual categorization~(FGVC) aims to achieve a refined classification of subclasses within a large class, where instances have similar appearance features and discrimination regions only exist locally. 
Previous research achieves this by utilizing large-scale annotated data and pre-trained deep models. However, when only a few-shot data is available, those methods may become less effective~\cite{li2023libfewshot,yan2023clip,xin2024few}.
To alleviate the pressure caused by the reduction in data quantity, 
relevant methods can be roughly divided into three categories including metric learning~\cite{snell2017prototypical,sung2018learning}, optimization, and data augmentation.
Specifically, metric learning uses predefined metrics to learn the deep representation of instances, and by calculating the distance or similarity between different images, they are divided into different categories.
In this process, pose-normalized representations are often used, which first locate the semantic parts in each image, and then describe the image by characterizing the appearance of each part~\cite{tang2020revisiting}.
Optimization methods often use transfer learning. Specifically, traditional deep learning is applied to adjust the source data, and then a simple classifier is trained to adjust the target data in a fixed representation~\cite{chen2019closer,tian2020rethinking}, or fine-tuned~\cite{dhillon2019baseline}.
Most data augmentation methods are based on an assumption that internal category variations caused by pose, background, or lighting conditions are shared between categories. Internal category variations can be modeled as low-level statistical information~\cite{yin2018feature} or pairwise transformations~\cite{hariharan2017low,schwartz2018delta}, and can be directly applied to new samples. 

Although these methods have been proven effective in general FSL tasks, the gains achieved in fine-grained datasets are minimal. 
For metric learning methods, The extracted features are difficult to form tight clusters for new classes because small changes in the feature space can be affected by small inter-class distances~\cite{xu2021variational,zhang2021prototype,lee2022task}.
For the optimization methods, fine-tuning FGVC images on large models is challenging because discriminative features often exist only locally. Limited samples often lead to pre-trained models struggling to comprehend instance details and perform effective recognition properly.
The data augmentation methods have shown significant promise, however, they also require careful design due to the risk of exacerbating the imbalance between discriminative and non-discriminative features~\cite{ye2023coping,hong2024improving}.

\subsection{Basic Principles of Diffusion Model}
Diffusion models are a type of latent variable models that include forward and reverse noise-injection process. During the forward process, noise is gradually added to the data, each step in the forward process is a Gaussian transition according to the following Markovian process
\begin{equation}
q\left( \boldsymbol{x}_t|\boldsymbol{x}_{t-1} \right) =\mathcal{N} \left( \sqrt{\alpha _t}\boldsymbol{x}_{t-1},\beta _t\mathbf{I} \right) ,\forall t\in \left\{ 1,...,T \right\},
\end{equation}
\begin{equation}
q\left( \boldsymbol{x}_{1:T}|\boldsymbol{x}_0 \right) =\prod_{t=0}^T{q\left( \boldsymbol{x}_t|\boldsymbol{x}_{t-1} \right)},
\end{equation}
where $T$ is the number of diffusion steps, The mean and variance of Gaussian noise are determined by $\beta_t$, $\alpha_{t}=1- \beta_t $. The reverse process is another Gaussian transition
\begin{equation}
    p_{\theta}\left( \boldsymbol{x}_{t-1}|\boldsymbol{x}_t \right) =\mathcal{N} \left( \boldsymbol{x}_{t-1}|\boldsymbol{\mu }_{\theta}\left( \boldsymbol{x}_t,t \right) ,\sigma _{\theta}^{2}\left( \boldsymbol{x}_t,t \right) \mathbf{I} \right),
\end{equation}
where the mean value $\boldsymbol{\mu }_{\theta}\left( \boldsymbol{x}_t,t \right) $ can be seen as the combination of  $\boldsymbol{x}_t$ and a noise prediction network $\epsilon _{\theta}\left( \boldsymbol{x}_t,t \right)$. The maximal likelihood estimation of the optimal mean is 
\begin{equation}
\tilde{\boldsymbol{\mu}}_{\theta}\left( \boldsymbol{x}_t,t \right) =\frac{1}{\sqrt{\alpha _t}}\left( \boldsymbol{x}_t-\frac{\beta _t}{\sqrt{1-\bar{\alpha}}}\mathbb{E} \left( \epsilon |\boldsymbol{x}_t \right) \right).
\end{equation}
To get the optimal mean, $\epsilon _{\theta}\left( \boldsymbol{x}_t,t \right)$ can be learned by a noise prediction objective
\begin{equation}
\min_{\theta} \mathbb{E} _{\boldsymbol{x}_0\sim q\left( \boldsymbol{x} \right) ,\epsilon \sim \mathcal{N} \left( 0,\mathbf{I} \right) ,t}\left\| \epsilon _{\theta}\left( \boldsymbol{x}_t,t \right) -\epsilon \right\| _{2}^{2}.
\end{equation}
However, in the context of conditional generation, the condition information $c$ should be considered during the training process of the noise prediction network
\begin{equation}
\min_{\theta} \mathbb{E} _{\boldsymbol{x}_0\sim q\left( \boldsymbol{x} \right) ,\epsilon \sim \mathcal{N} \left( 0,\mathbf{I} \right) ,t,c}\left\| \epsilon _{\theta}\left( \boldsymbol{x}_t,t,c \right) -\epsilon \right\| _{2}^{2}.
\end{equation}

This process is typically implemented using the U-Net~\cite{ronneberger2015u} architecture. In order to leverage information from different modalities, recent studies have also incorporated Transformer's self-attention~\cite{dosovitskiy2020image,liu2019mtfh,liu2021fddh,peng2023towards} modules~(including a self-attention layer, a cross-attention layer, and a fully connected feed-forward network) for feature alignment~\cite{baykal2023protodiffusion}.
Specifically, the attention layer operates on queries $ \boldsymbol Q \in \mathbb R^{n\times d_{k}}$, and key-value pairs $ \boldsymbol K \in \mathbb R^{m\times d_{k}}$, $ \boldsymbol V \in \mathbb R^{m\times d_{v}}$
\begin{equation}
A(\boldsymbol Q,\boldsymbol K,\boldsymbol V) = \text {softmax} (\frac{ \boldsymbol Q \boldsymbol K^{T}}{\sqrt{d_{k}}}) \boldsymbol V,
\end{equation}
where $n$ is the number of queries, $m$ is the number of key-value pairs $d_{k}$ is dimension of key, $d_{v}$ is the dimension of value. In the self-attention layer, $\boldsymbol x \in \mathbb R^{n\times d_{x}}$ is the only input. In the crosss attention layer of the conditioned diffusion model, there are two inputs $\boldsymbol x \in \mathbb R^{n\times d_{x}}$ and $\boldsymbol c \in \mathbb R^{m\times d_{c}}$,
where $\boldsymbol x$ is the output from the prior block and $\boldsymbol c$ represents the condition information.
However, diffusion models cannot be used directly for data augmentation of fine-grained images due to fine-grained labels possess a certain level of expertise, and pre-trained models have difficulty in understanding the mapping between labels and semantics (see more details in Section IV). 

\begin{figure*}[hbpt]
  \begin{center}
  \includegraphics[width=0.99\textwidth]{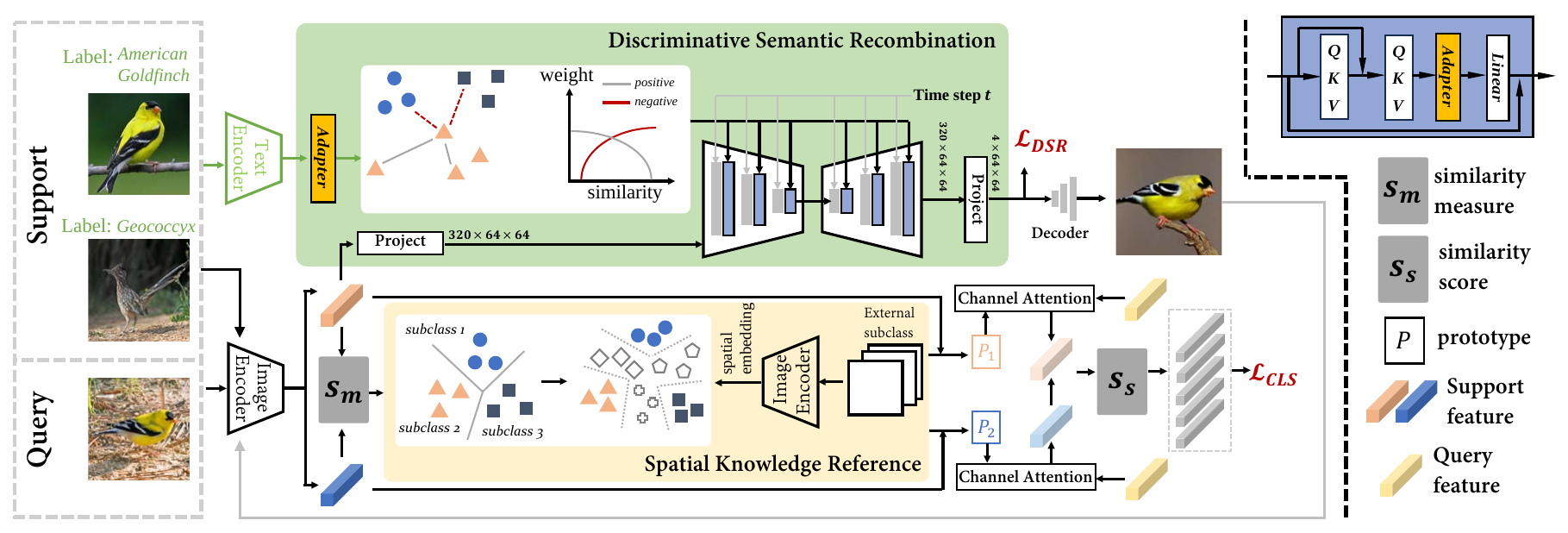}
  \end{center}
	\setlength{\abovecaptionskip}{-0.1cm} 
   \caption{Overview of the proposed method. Our framework first uses DSR to constrain similarity relations from the labels, thereby enhancing the model's understanding of instance-specific features. Then, during the classification process, we introduce instance features from different datasets for comparative reference, ensuring that the learned features possess a stronger representational capacity and robustness.}
\vspace{-0.4cm} 
  \label{overview}
\end{figure*}
\section{Method}
In this section, we describe our DRDM. As shown in Figure~\ref{overview}, it consists of two
core components, including discriminative semantic recombination, and spatial knowledge reference module.

\subsection{Notation}
In the few-shot FGVC tasks, the dataset is divided into meta-training set $D_{base}=\left\{\left(x_i,y_i\right),y_i\in C_{base}\right\}$ and meta-testing set $D_{novel}=\left\{\left(x_k,y_k\right),y_k\in C_{novel}\right\}$, where $C_{base}$ and $C_{novel}$ represent base and novel classes respectively, and $C_{base}\cap C_{novel}=\phi$. Here, $x_k$, and $y_k$ denote input image and class name respectively.
Furthermore, during the training and testing phases of few-shot FGVC tasks, they are typically composed of distinct episodes. Each episode includes a labeled support set $S=\left\{\left(x_k,y_k,\right)\right\}_{k=1}^{N\times K}$ and an unlabeled query set $Q=\left\{\left(x_k,y_k\right)\right\}_{k=1}^{N\times U}$. In these sets, $N$ signifies the number of randomly selected classes, and $S$ and $Q$ share the same class. $K$ and $U$ represent the quantities of labeled and unlabeled samples respectively, while ensuring $S\cap Q=\phi$.
For ease of reference, we have compiled some important symbols and definitions in Table~\ref{symbol}.

\begin{table}[hbtp]\small
	\centering
	\caption{Partial symbols and definition explanations.}
	\begin{tabular}{l|l}
		\hline
		\textbf{Symbol}   & \textbf{Definition} \\  
		\hline
		$D_{base}$,$D_{novel}$   &  Meta-training and meta-test set \\ 
        $D_{extra}$       & Additional meta-training set \\
		$F_{i,j}^S$,$F_{i,j}^Q$   & Computed feature of support and query set  \\
        $F_{i,j}^E$  & Computed feature of extra set  \\
		$F_{i}^{P}$             & Prototype representative of $i$-th class  \\  
		$R_{i,c}^{intra}$,$R_{i,c}^{inter}$ & Intra-class and inter-class representation score \\ 
		$w_{i}^{intra}$,$w_{i}^{inter}$ & Intra-class and inter-class attention weights \\ 
		$w_i$ & Channel attention weights \\
		$\mathcal{L}_{MS}$ & Multi-Similarity loss \\
		\hline
	\end{tabular}
	\label{symbol}
\end{table}

\vspace{-0.4cm} 

\subsection{Discriminative Semantic Recombination~(DSR)}
As mentioned above, the implicit similarity descriptions in the textual modality encompass the fundamental features of the subclasses. We argue that transferring this part of similarity knowledge from the text space to the image space would aid in augmenting the data with more intricate details, thereby generating features with stronger representational capabilities.
To achieve this, we first conduct similarity measurements in the text space. 
In this process, the model needs to establish a connection between fine-grained labels and instances, which becomes challenging under the few-shot paradigm. Utilizing too few instances to fine-tune a large model is not sufficient for the model to acquire enough discriminative knowledge and may lead to severe overfitting.
Recently, a novel fine-tuning paradigm has emerged with the use of Adapter methods (e.g., AdapterFusion~\cite{pfeiffer2020adapterfusion}, AdapterDrop~\cite{ruckle2020adapterdrop}, and K-Adapter~\cite{wang2020k}). This paradigm involves adding Adapter modules to certain layers of a pre-trained model and freezing the pre-trained backbone during fine-tuning. The Adapter modules are responsible for learning specific downstream task knowledge, thereby avoiding the issues of full model fine-tuning and catastrophic forgetting. 

Inspired by this, in our approach, we introduce Adapter knowledge layers into the process of interconnecting textual and visual features of the diffusion model. With only a few parameters specifically designed for the fine-grained task, these Adapter knowledge layers store instance-specific knowledge, thereby mitigating the overfitting issues that may arise from fine-tuning.
We introduced an adapter module into the U-Net architecture after the cross-attention layer for visual features. At the same time, an adapter was added after the pre-trained text encoder to transfer the semantic understanding.
Defining the prompt as $V$, expressed as ``a photo of a [class label]". the formulation becomes $F_C=\Phi_{AD}\left(Z\left(V\right)\right)$, where $Z$ represents the encoder used for the prompt input, such as CLIP~\cite{radford2021learning}. $\Phi_{AD}$ signifies the added adapter module.
The loss function for the CLIP branch is denoted as $\mathcal{L}_C=\mathcal{L}_{MS}\left(F_C\right)$ where $\mathcal{L}_{MS}$ refers to the MS loss~\cite{wang2019multi}, computed as follows:

\begin{equation}
\begin{aligned}
\mathcal{L}_{MS}= &\frac{1}{m}\sum_{i=1}^{m}\left\{\frac{1}{2}log\left[1+\sum_{k\in\mathcal{P}_i} e^{-2\left(S_{ik}-\frac{1}{2}\right)}\right] \right. \\ & \left. +\frac{1}{40}log\left[1+\sum_{k\in\mathcal{N}_i} e^{40\left(S_{ik}-\frac{1}{2}\right)}\right]\right\},
\end{aligned}
\end{equation}
The reconstruction loss of Stable Diffusion ($\mathcal{L}_{SD}$) is defined as follows:
\begin{equation}
\mathcal{L}_{SD}=\mathbb{E}_{t,x_0,c,\epsilon}||\epsilon_\theta\left(x_t,t,c\right)-\epsilon||^2_{2}.
\end{equation}

The overall loss function in DSR is given by:
\begin{equation}
\mathcal{L} _{DSR}=\mathcal{L}_{SD}+\alpha \mathcal{L}_C.
\end{equation}
where $\mathcal{L}_C$ is the CLIP branch loss, and $\alpha$ is a hyperparameter controlling the trade-off between the reconstruction loss and the CLIP branch loss.

\subsection{Spatial Knowledge Reference~(SKR)}
One of the challenges in few-shot FGVC is the limited number of samples, which results in the subclass features becoming highly scattered when mapped to a high-dimensional space. Consequently, the classifier struggles to identify clear and distinct subclass feature boundaries in the high-dimensional space, significantly affecting the accuracy and performance of FGVC tasks.
We argue that leveraging the knowledge from other datasets as a reference can significantly enhance the model's understanding of similar instances. 

\begin{figure}[hbpt]
  \begin{center}
  \includegraphics[width=0.49\textwidth]{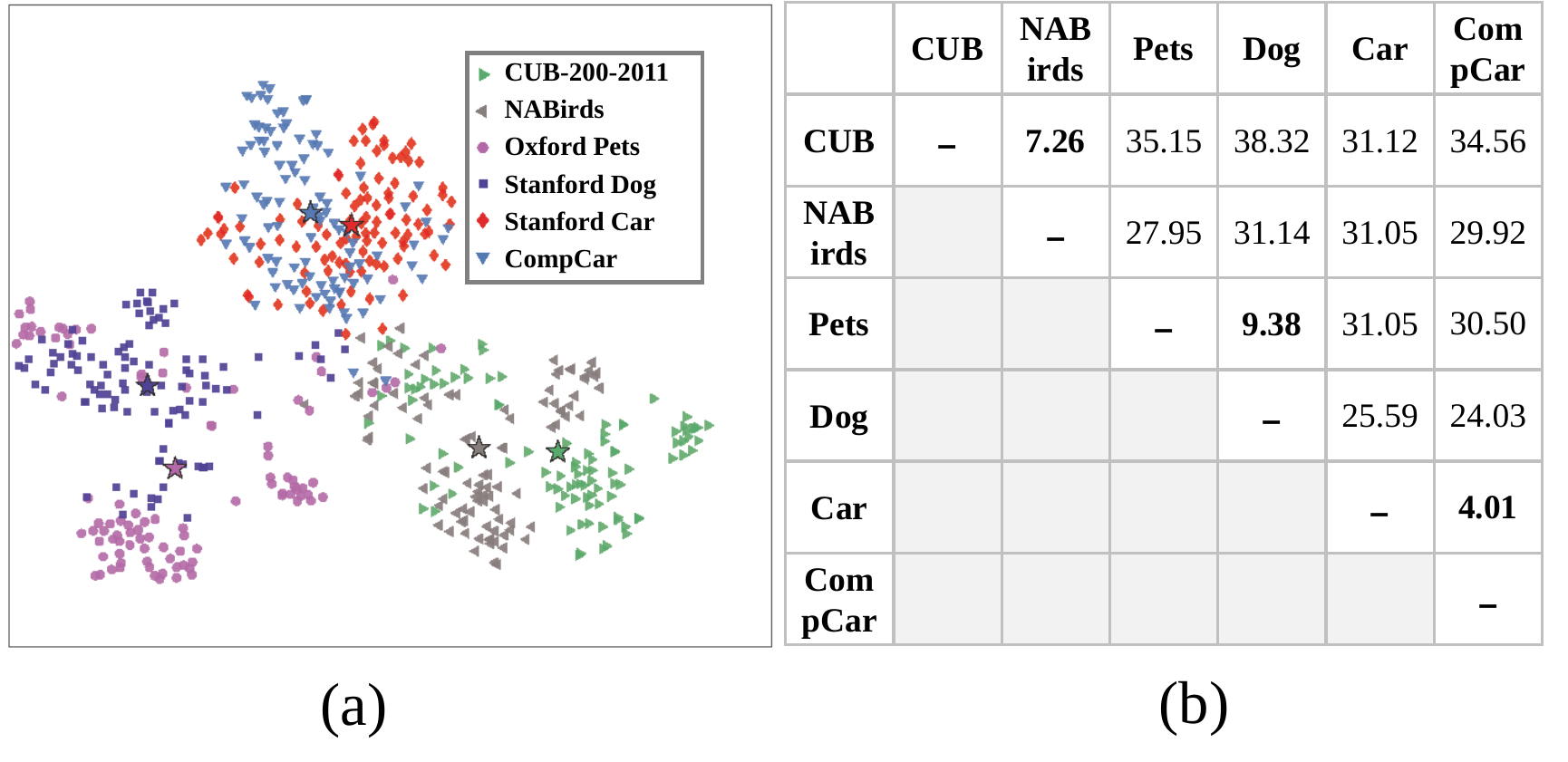}
  \end{center}
	\setlength{\abovecaptionskip}{-0.1cm} 
   \caption{Analysis of dataset feature distributions. All features are extracted using a pre-trained ResNet-50. (a) Qualitative analysis, where points of different colors represent different datasets. (b) Quantitative analysis, which calculates the distances between the centers of each dataset.}
\vspace{-0.1cm} 
  \label{add1}
\end{figure}

One reason to consider is that data from a similar class tends to be closer in high-dimensional space. As shown in Figure~\ref{add1}.
It can be observed that CUB exhibits significant overlap in feature distribution with NABirds, Stanford Dogs, and Oxford-Pet, as well as Stanford Cars and CompCar, with the distribution distances noticeably closer compared to other classes of data. Referencing similar datasets can help the model understand the subtle differences between instances within similar features, encouraging the model to find clear and distinct boundaries for fine-grained features in the high-dimensional space.
However, existing research on few-shot FGVC fails to utilize foundational knowledge from similar data, as the datasets' training processes are disjointed. Additionally, directly applying knowledge from other datasets not only fails to improve the model's performance but also leads to severe learning degradation.

To address these issues, we have designed a knowledge reference module, which aims to effectively incorporate knowledge from other datasets in a coherent manner during the training process. This module allows the model to benefit from the shared knowledge of similar data, leading to improved performance in few-shot FGVC tasks.

During the training phase of base class, in addition to the support set and query set, we also introduced additional set $E=\left\{\left(x_k,y_k\right)\right\}_{k=1}^{W\times K}\subseteq D_{extra}$, where $W$ represents the number of additional dataset classes added, and $D_{extra}=\left\{\left(x_i,y_i\right),y_i\in C_{extra}\right\}$, satisfying $C_{base}\cap C_{extra}=\phi$ and $C_{novel}\cap C_{extra}=\phi$.
The training framework involves a network feature extractor, denoted as $f\left( \cdot |\theta \right)$, which is responsible for computing features for different sets.
The prototype representative of each class is expressed as: $F_{i}^{P}=\frac{1}{K}\sum_{j=1}^K{F_{i,j}^{S}}$,
where $F_{i,j}^{S}$ denotes the $j$-th feature in $i$-th class. Inspired by TDM~\cite{lee2022task}, we applied channel attention to the feature of both the support set and query set. Firstly, the intra-class representation score across channel dimensions is defined by:
\begin{equation}
R_{i,c}^{intra}=\frac{1}{H\times W}\left\| F_{i,c}^{P}-M_{i}^{P} \right\| ^2,
\end{equation}
where $H$ and $W$ represent the width and height of the feature and $M_{i}^{P}\in R^{H\times W}$ represents the mean prototype feature across channel. Secondly, the inter-class representation score is calculated by:
\begin{equation}
	R_{i,c}^{inter}=\underset{i\in C,j\in C,i\ne j}{\min}\frac{1}{H\times W}\left\| F_{i,c}^{P}-M_{j}^{P} \right\| ^2 .
\end{equation}

Subsequently, the obtained intra-class representation score and inter-class representation score are passed through the fully connected network to obtain the attention weights for different channels of the $i$-th class:
\begin{equation}
	w_{i}^{intra}=f_{intra}\left( R_{i}^{intra} \right),
\end{equation}
\begin{equation}
	w_{i}^{inter}=f_{inter}\left( R_{i}^{inter} \right),
\end{equation}
where $f_{intra}\left( \cdot \right) $ and $f_{inter}\left( \cdot \right) $ denote the fully connected networks for intra-class scores and inter-class scores, respectively. The final channel attention weight is given by:
\begin{equation}
	w_i=( {w_{i}^{intra}+w_{i}^{inter}} )/ {2},
\end{equation}
We apply channel attention weight to the prototype representation and query set of each class:
\begin{equation}
	G_{i}^{S}=w_i\odot F_{i}^{P},
\end{equation}
\begin{equation}
	G_{i,j}^{Q}=w_i\odot F_{i,j}^{Q}.
\end{equation}
Finally, according to ProtoNet~\cite{snell2017prototypical}, the inference results of query set are given by 
\begin{equation}
	p\left( \left. y=i \right|x \right) =\frac{\exp \left( -dist\left( G_{i}^{S},G_{i}^{Q} \right) \right)}{\sum\nolimits_{j=1}^N{\exp \left( -dist\left( G_{j}^{S},G_{j}^{Q} \right) \right)}},
\end{equation}
where $dist\left( \cdot \right)$ denotes similarity distance measure between features.
The computation of SKR gives rise to the corresponding loss term, denoted as $\mathcal{L}_{SKR}=\mathcal{L}_{MS}\left(Cat\left(F^S, F^E\right)\right)$ with $Cat\left(\cdot\right)$ representing the concatenation operation.
The overall loss function for the final classification network is expressed as:

\begin{equation}
\mathcal{L} _{CLS}=-\frac{1}{N\times U}\sum_{k=1}^{N\times U}{\left( \boldsymbol{y}_{k}^{T}\log \left( \boldsymbol{p}_k \right) \right)}+\beta \mathcal{L} _{SKR},
\end{equation}
where $\boldsymbol{y}_k$ stands for the one-hot vector and $\boldsymbol{p}_k$ for predicted probability, and $\beta$ is a hyperparameter controlling the balance between the classification cross-entropy loss and the SKR loss.

\section{Experiments}
In this section, we extensively evaluated the performance of our approach.
We compared the performance of our approach with the latest state-of-the-art~(SOTA) methods on each network architecture.
The experimental settings, implementation details, and results for diverse tasks are described below.

\subsection{Datasets and Experimental Setup}
Experiments are conducted on three widely used datasets.
All datasets provide fixed train and test splits.
The details are summarized in Table~\ref{dataset}. 
\renewcommand{\arraystretch}{1.0}
\begin{table}[hbtp]\normalsize
    \centering
    \caption{The splits of datasets. While \textit{C}$_{all}$ is the number of total subclasses, \textit{C}$_{train}$, \textit{C}$_{val}$, \textit{C}$_{test}$ are the number of training, validation, and test subclasses, respectively. The classes of subsets are disjoint.}
    \begin{tabular}{l|c|c|c|c}
        \hline
          \textbf{Dataset}   & \textit{C}$_{all}$  & \textit{C}$_{train}$  & \textit{C}$_{val}$  & \textit{C}$_{test}$ \\  
        \hline
CUB-200-2011\cite{wah2011caltech}               & 200    & 100  & 50  & 50    \\  
Stanford Dogs\cite{khosla2011novel}             & 120    & 60   & 30  & 30    \\  
Stanford Cars\cite{krause20133d}	             & 196    & 130  & 17  & 49    \\
\hline
    \end{tabular}
    \label{dataset}
\end{table}

For the CUB dataset, our data split is the same with~\cite{wertheimer2021few}.
Regarding the Cars dataset, we adhere to the same data split with~\cite{li2019revisiting}.
As for the Dogs dataset, it comprises 90 subclasses designated for training and validation, along with an additional 30 subclasses for testing.
To achieve effective spatial knowledge referencing, we employed the NABirds~\cite{van2015building}, Oxford-IIIT Pet~\cite{parkhi2012cats}, and CompCars~\cite{yang2015large} datasets as supplementary knowledge sources for the CUB, Dogs, and Cars datasets, respectively.
In the experiments, ResNet~\cite{he2016deep} is pre-trained on ImageNet as the backbone, and all the input images are cropped to $84\times84$. 
The model is trained with the stochastic gradient descent~(SGD) and momentum of 0.9 for all datasets.
The initial learning rate of the main branch was set to 0.001 and 0.01 for the rest layers.
Our implementation is based on PyTorch with an NVIDIA Geforce GTX 3090Ti GPU. In the N-way K-shot scenario, we carried out few-shot classification on 10,000 randomly sampled episodes, each containing 16 queries per class. We present the average classification accuracy along with 95\% confidence intervals, as in ~\cite{lee2022task}.

\subsection{Model Configuration}
Model configuration experiments are conducted to verify the validity of the individual component and to determine the hyperparameters. 

\textbf{Multi-Similarity Loss ($\boldsymbol{\alpha}$):}
To verify the effectiveness of MS loss and investigate the influence of the parameter $\alpha$, extensive experiments are carried out on the three datasets, and the results are presented in Table~\ref{MS}. 

\begin{table}[hbpt]\small
 \begin{center}
\caption{Experimental results using varied $\alpha$. “w/o” means learning without MS loss. The best performance is indicated in bold.}
\label{MS}
 \begin{tabular}{c|ccccc|c}  
 \bottomrule 
  $\alpha$  & 0.1  & 0.3  & 0.5  & 0.7  & 0.9 & w/o\\
 \hline
CUB      & 88.09  & 88.40  & \bf 88.53  & 88.35  & 88.05  &88.14\\
Dogs      & 71.38  & 72.11  & \bf 72.28  & 71.93  & 72.51  &72.04\\
Car      & 80.58  & 80.83  & \bf 81.03  & 80.80  & 80.63  &80.38\\
  \hline  
\end{tabular} \end{center}
\vspace{-0.8cm}  
\end{table}

When $\alpha$ was set properly (e.g., $\alpha \in [0.3~0.7]$), the MS loss could effectively embed the information contained in the label into the image space, to some extent facilitating the model's comprehension of similar instances.
However, an increment in $\alpha$ beyond a certain point led to a slight decline in our model’s performance. 
One possible reason is that the model overly relies on the relationships among labels within the few-shot training paradigm, potentially resulting in overfitting.
This suggests that $\alpha=0.5$ could be a reliable choice for DRDM.

\begin{table*}[hbtp]  
\caption{Few-shot classification accuracy on the CUB, Stanford Dogs, and Stanford Cars dataset. All experiments are from a 5-way classification with the same backbone network (ResNet12). The best performance is indicated in bold.}
  \centering  
  \fontsize{1.5}{2}\small
  \begin{threeparttable}  
 \begin{tabular}{c|cc|cc|cc}  
\toprule  
 \multirow{2}{*}{\textbf{Method}}  &\multicolumn{2}{c|}{\bf CUB-200-2011}   &\multicolumn{2}{c|}{\bf Stanford Dogs} &\multicolumn{2}{c}{\bf Stanford Cars}\\
    &\textbf{1-shot}   & \textbf{5-shot}  &\textbf{1-shot}   & \textbf{5-shot}  &\textbf{1-shot}   & \textbf{5-shot}  \\ 
\midrule   
MTL~\cite{sun2019meta}~(CVPR@2019)                   &73.31$\pm$0.92  &82.29$\pm$0.51  &54.96$\pm$1.03 &68.76$\pm$0.65  &-  &- \\
MetaOptNet~\cite{lee2019meta}~(CVPR@2019)        &75.15$\pm$0.46  &87.09$\pm$0.30  &65.48$\pm$0.49 &79.39$\pm$0.25  &-  &- \\
S2M2~\cite{mangla2020charting}~(WACV@2020)      &71.43$\pm$0.28  & 85.55$\pm$0.52  &-&- &-&-\\
Neg-Cosine~\cite{liu2020negative}~(ECCV@2020)     &72.66$\pm$0.85  & 89.40$\pm$0.43  &-&- &-&- \\
A2~\cite{afrasiyabi2020associative}~(ECCV@2020)    &74.22$\pm$1.09  & 88.65$\pm$0.55  &-&- &-&-\\

MattML~\cite{zhu2020multi}~(IJCAI@2020)      &66.29$\pm$0.56 &80.34$\pm$0.30 &54.84$\pm$0.53 &71.34$\pm$0.38 &66.11$\pm$0.54 &82.80$\pm$0.28\\
BSNet(R\&C)~\cite{li2020bsnet}~(TIP@2021)    &65.89$\pm$1.00  &80.99$\pm$0.63   &51.06$\pm$0.94& 68.60$\pm$0.73   &54.12$\pm$0.96 & 73.47$\pm$0.75\\
VFD*~\cite{xu2021variational}~(ICCV@2021)    &79.12$\pm$0.83   &91.48$\pm$0.39  & 57.04$\pm$0.89  & 72.95$\pm$0.70  &-  &- \\
APF~\cite{tang2022learning}~(PR@2022)        &78.73$\pm$0.84  &89.77$\pm$0.47  &60.89$\pm$0.98 & 78.14$\pm$0.62   &78.14$\pm$0.84 & 87.42$\pm$0.57\\
TDM~\cite{lee2022task}~(CVPR@2022)           & 84.36$\pm$0.19  &93.37$\pm$0.10  &57.32$\pm$0.22        & 75.26$\pm$0.16        &67.10$\pm$0.22  &86.05$\pm$0.12 \\
CFMA~\cite{li2022coarse}~(IS@2022) &74.68$\pm$1.38	&90.91$\pm$0.94  &-  &- &- &- \\
QGN~\cite{munjal2023query}~(PR@2023)   & 83.82$\pm$0.00	&91.22$\pm$0.00  &-  &- &- &-\\
T2L~\cite{sun2023t2l}~(KBS@2023)   &71.04$\pm$1.21 &83.44$\pm$0.94 &52.12$\pm$1.14 &70.83$\pm$1.09 &56.80$\pm$1.23 &74.10$\pm$1.65 \\
TasNet~\cite{pan2023few}~(PR@2023) &83.89$\pm$0.69	 &91.35$\pm$0.53 &-  &- &- &- \\

\midrule  
Ours                                                                       &\bf 89.99$\pm$0.14  &\bf94.63$\pm$0.10   &\bf72.68$\pm$0.17&\bf80.43$\pm$0.12 &\bf81.53$\pm$0.15&\bf90.03$\pm$0.12\\
\bottomrule
    \end{tabular}  
    \end{threeparttable}  
 \label{last_3_years}
\vspace{-0.4cm} 
\end{table*}

\begin{figure}[hbpt]
  \begin{center}
  \includegraphics[width=0.49\textwidth]{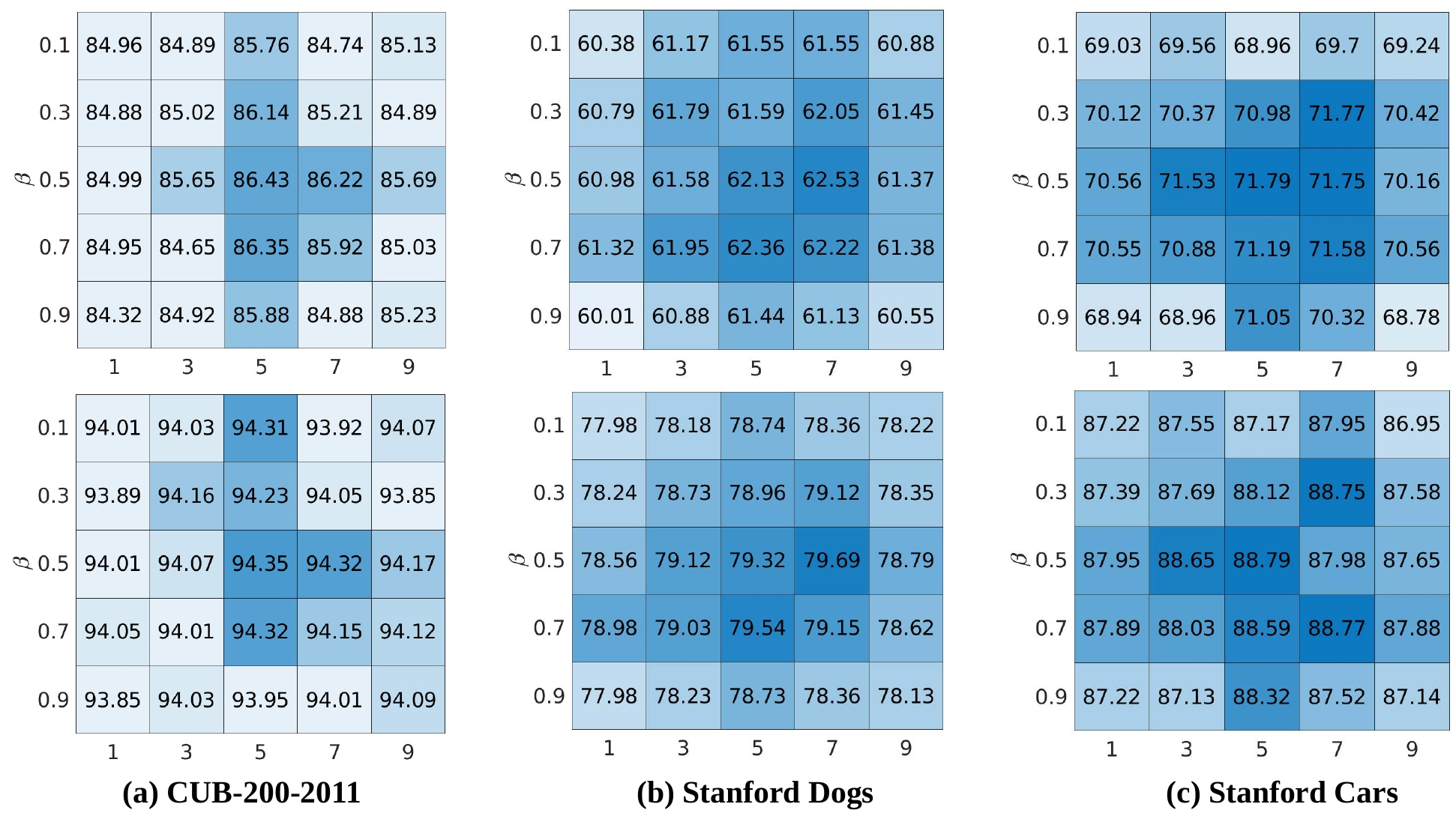}
  \end{center}
	\setlength{\abovecaptionskip}{-0.1cm} 
   \caption{The impact of different N and $\beta$ on the learning process. The horizontal and vertical axes represent the number of introduced subclasses and the selection of $\beta$ values, respectively. The first and second rows represent the results for the 1-shot and 5-shot scenarios, respectively, with different columns showing the results for different datasets. The color intensity is used to visualize the level of accuracy, where darker shades indicate higher accuracy.}
\vspace{-0.2cm} 
  \label{parameter}
\end{figure}

\textbf{Spatial Knowledge Reference ($\boldsymbol{\beta}$):}
To investigate the influence of SKR, we measured the impact of introducing the number of subclasses (N) with different $\beta$ values for both 1-shot and 5-shot scenarios on three datasets, the results are shown in Figure~\ref{parameter}.

As can be seen, in the CUB dataset, the model’s accuracy exhibited a positive trend after incorporating subclasses as reference knowledge. However, when the number exceeded five, a slight reduction in accuracy was observed. This might be attributed to the challenge of the model in comprehending target subclasses under the setting of few-shot problems when too many subclasses were introduced. 
Besides, we observed that increasing $\beta$ leads to higher accuracy. However, the performance modestly decreased as the balance parameter $\beta$ was increased from 0.5 to 0.7, suggesting that when $\beta$ was equal to 0.5, the model was able to leverage a sufficient amount of knowledge.
Similar results can be observed in the other two datasets. Therefore, we still chose $N = 5$ and $\beta = 0.5$ for all subsequent experiments, as it ideally balances computational complexity and accuracy.

\subsection{Performance Evaluation}

\begin{figure*}[hbpt]
  \begin{center}
  \includegraphics[width=0.98\textwidth]{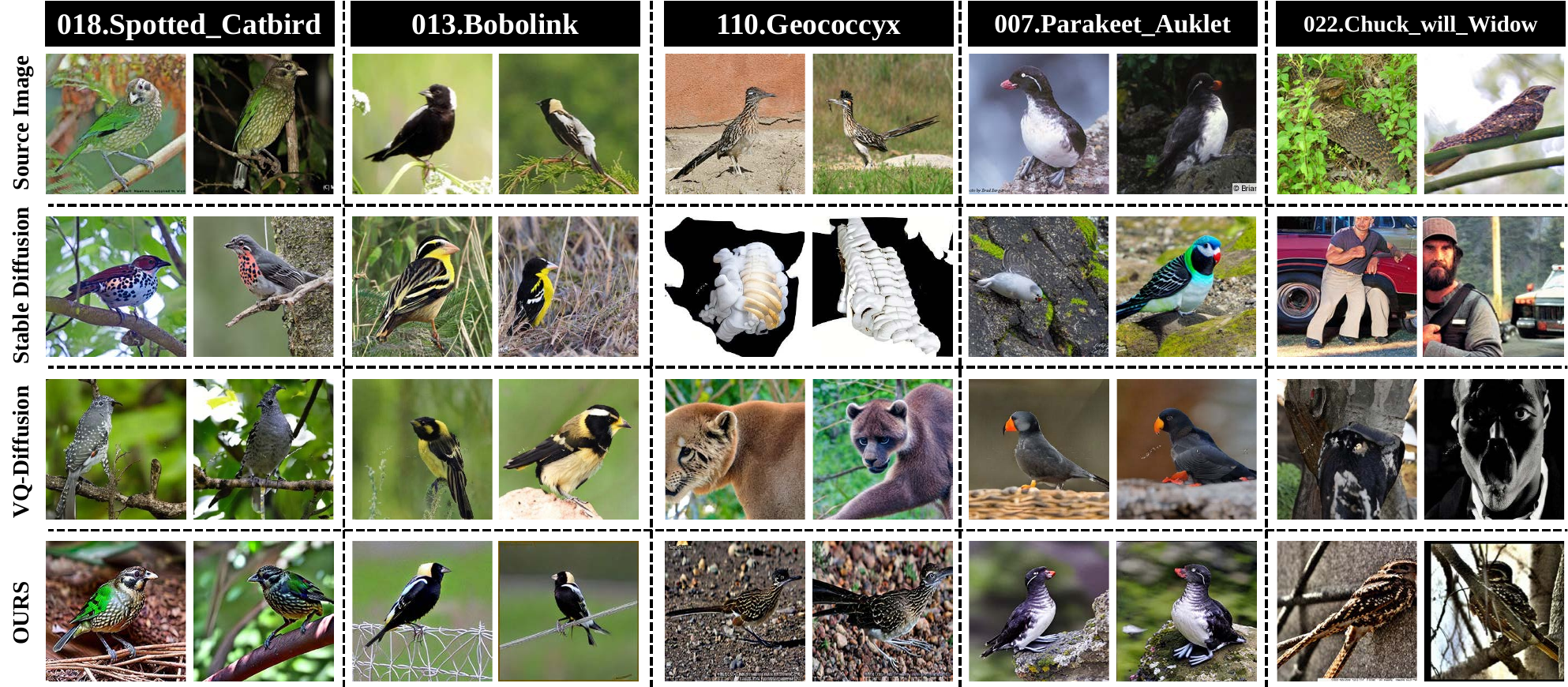}
  \end{center}
	\setlength{\abovecaptionskip}{-0.1cm} 
   \caption{Comparison of fine-grained image data augmented by Diffusion-Based Models. The first row depicts the source images, the second and third rows demonstrate the results generated by part of existing methods, and the last row exhibits our results.}
\vspace{-0.1cm} 
  \label{VisualizationofSD}
\end{figure*}

The experimental results and analysis of DRDM compared with recent SOTA methods on three datasets are presented in Table~\ref{last_3_years}.
It was observed that on the CUB dataset, the TDM~\cite{lee2022task} method demonstrated SOTA performance, which can be credited to its channel attention mechanism. This mechanism produced a support weight to represent the channel-wise discriminative power for each subclass. Benefiting from the embedding and learning of textual feature space relationships, our approach achieved performance improvements of 5.63\% and 1.26\% in the 1-shot and 5-shot settings, respectively.
On the Dogs dataset, the SOTA performance was achieved by MetaOptNet~~\cite{lee2019meta}. Similarly, this approach overlooks the potential structural relationships within the labels. Consequently, in comparison, our method achieved performance improvements of 7.2\% and 1.04\% in the 1-shot and 5-shot settings, respectively.
On the Cars dataset, the SOTA performance is achieved by APF~\cite{tang2022learning}. In comparison, our method achieved performance improvements of 3.39\% and 2.61\% in the 1/5-shot settings, respectively.
These results provide compelling evidence for the effectiveness of DRDM.


To validate the unique advantage of our proposed algorithm in fine-grained data augmentation,
We also compared it with data augmentation methods based on the diffusion model.
Instances augmented by different diffusion models are demonstrated in Figure~\ref{VisualizationofSD}.
\begin{table*}[hbtp]  
\caption{Comparison of data augmentation methods based on diffusion model. The best performance is indicated in bold.}
  \centering  
  \fontsize{1.5}{2}\small
  \begin{threeparttable}  
 \begin{tabular}{c|cc|cc|cc}  
\toprule  
 \multirow{2}{*}{\textbf{Method}}  &\multicolumn{2}{c|}{\bf CUB-200-2011}   &\multicolumn{2}{c|}{\bf Stanford Dogs} &\multicolumn{2}{c}{\bf Stanford Cars}\\
    &\textbf{1-shot}   & \textbf{5-shot}  &\textbf{1-shot}   & \textbf{5-shot}  &\textbf{1-shot}   & \textbf{5-shot}  \\ 
\midrule   
LDM~\cite{rombach2022high}~(CVPR@2022)   &82.82$\pm$0.18  & 90.12$\pm$0.12  &58.77$\pm$0.18	&70.48$\pm$0.16  &73.70$\pm$0.18  & 86.50$\pm$0.12   	\\

VQ-Diffusion~\cite{gu2022vector}~(CVPR@2022)   &81.54$\pm$0.19  & 89.96$\pm$0.12  &60.99$\pm$0.19	&70.60$\pm$0.16  &62.02$\pm$0.21  & 85.12$\pm$0.13   	\\

\midrule  
Ours    &\bf 89.99$\pm$0.14  &\bf94.63$\pm$0.10   &\bf72.68$\pm$0.17&\bf80.43$\pm$0.12 &\bf81.53$\pm$0.15&\bf90.03$\pm$0.12\\
\bottomrule
    \end{tabular}  
    \end{threeparttable}  
 \label{cwithSDiffusion}
\vspace{-0.2cm} 
\end{table*}  
It was observed that, while the data augmented by existing diffusion-based models could capture the basic outline and features of the instances~(e.g., \textit{Spotted Catbird}, \textit{Bobolink}, and \textit{Parakeet Auklet}), the limited of training samples makes it challenging for these models to capture finer details. For instance, the feathers of \textit{Spotted Catbird} are green, but the augmented data shows grey feathers. Moreover, the labels of fine-grained images have a certain level of expertise, making it difficult for large models to establish a clear mapping between labels and semantic features during pre-training. As a result, the augmented images may not correctly represent the instances~(e.g., the augmented data of \textit{Geococcyx} and \textit{Chuck-will-widow}). This leads to significant feature contamination, impeding model learning in a few-shot setting.

This conclusion is proved quantitatively in Table~\ref{cwithSDiffusion}. To ensure a fair comparison, all results were obtained using the same framework.
It can be observed that utilizing the SOTA diffusion-based model for data augmentation has led to a decrease in 1-shot accuracy compared to prior research, with declines of 1.54\%, 4.49\%, and 4.44\% on the three datasets~(contrasting the results in Table~\ref{last_3_years}), respectively. This reveals the prevalent presence of feature contamination and its impact on few-shot FGVC learning.

\begin{table*}[hbtp]\small
    \centering
\caption{Ablation studies of the DRDM on three datasets. The best performance is indicated in bold.}
    \begin{tabular}{c|cc|cc|cc|cc}  
        \hline
        \multirow{2}{*}{\textbf{Dataset}}  &\multicolumn{2}{c|}{\textbf{Framework}}  &\multicolumn{2}{c|}{\textbf{Framework + SKR}} &\multicolumn{2}{c|}{\textbf{Framework + DSR}} &\multicolumn{2}{c}{\textbf{DRDM}}\\
    &\textbf{1-shot}   & \textbf{5-shot}  &\textbf{1-shot}   & \textbf{5-shot}  &\textbf{1-shot}   & \textbf{5-shot}  &\textbf{1-shot}   & \textbf{5-shot} \\ 
\hline
CUB   &84.18$\pm$0.19  & 93.44 $\pm$0.10 &86.43$\pm$0.16  & 93.24$\pm$0.10  &88.14$\pm$0.15  & 93.62$\pm$0.10 &\bf89.99$\pm$0.14  &\bf94.63$\pm$0.10\\
Dogs  &61.44$\pm$0.22  &78.73$\pm$0.15	 &62.53$\pm$0.23	&79.69$\pm$0.15	&72.14$\pm$0.18	&79.86$\pm$0.15&\bf72.68$\pm$0.17&\bf80.43$\pm$0.12\\
Cars   &67.10$\pm$0.22  & 86.05$\pm$0.12  &71.79$\pm$0.21  & 88.79$\pm$0.12  &80.58$\pm$0.17  & 89.52$\pm$0.11 &\bf81.53$\pm$0.15&\bf90.03$\pm$0.12\\
        \hline
    \end{tabular}
\vspace{-0.2cm}
    \label{Ablation_Study}
\end{table*}

\begin{figure*}[hbpt]
  \begin{center}
  \includegraphics[width=0.95\textwidth]{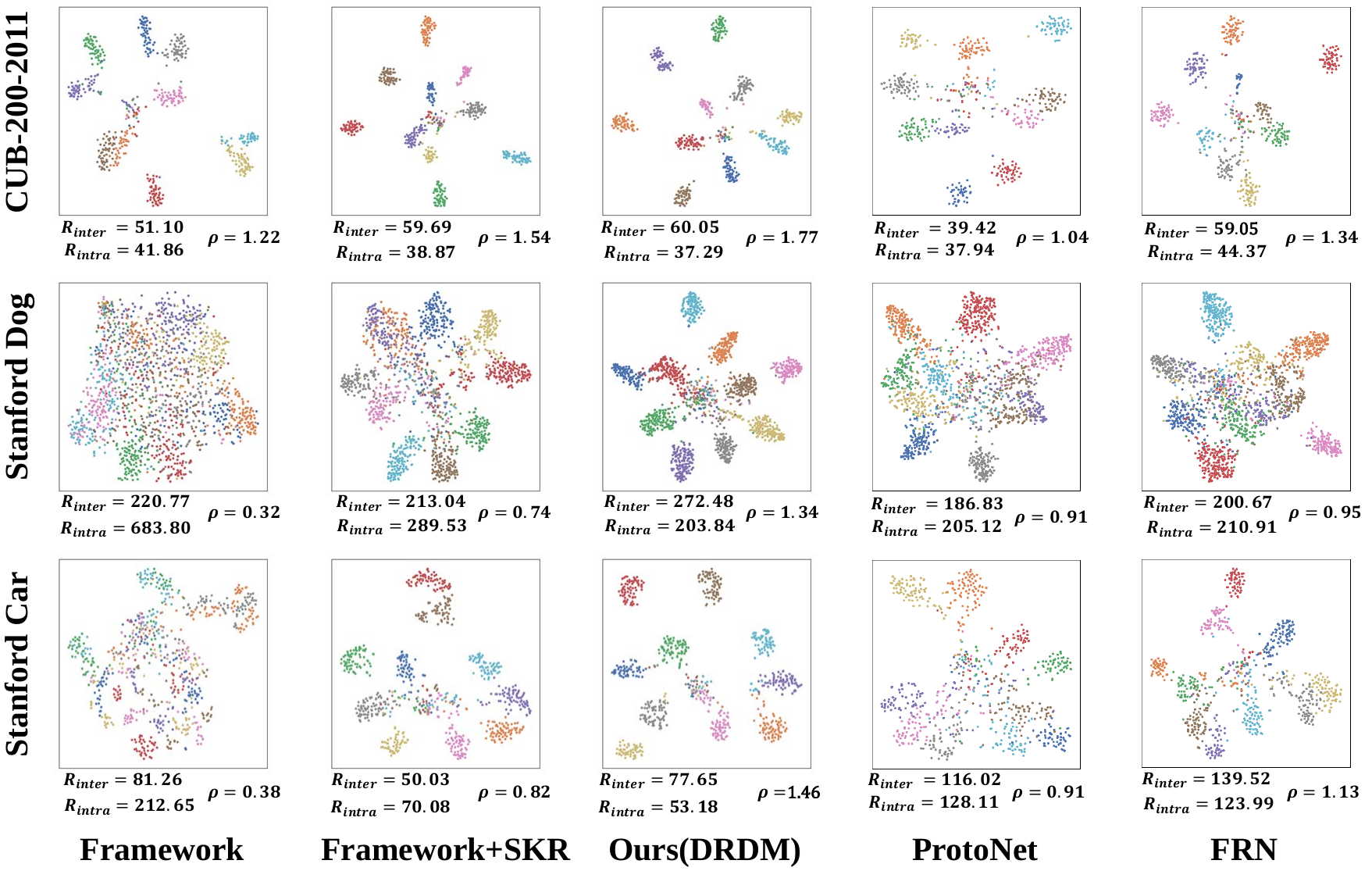}
  \end{center}
	\setlength{\abovecaptionskip}{-0.1cm} 
   \caption{The feature visualization using t-SNE~\cite{van2008visualizing}. Each row displays the results of different datasets, while each column represents the distribution of features at different stages. Each dot denotes the feature of a sample and different colors represent different categories. $R_{inter}$ and $R_{intra}$ indicate the degree of compactness of inter-class and intra-class, respectively. $\rho$ means space utilization~\cite{tian2019sosnet}.}
\vspace{-0.1cm} 
  \label{TSNE}
\end{figure*}

\subsection{Ablation Study}
To evaluate the proposed DRDM, an ablation study was conducted. 
``Framework" refers to using a structure designed without any strategy. 
On this basis, the influence of DSR and SKR strategies on learning was explored. The experimental results are presented in Table~\ref{Ablation_Study}. 

Taking CUB as an example, it can be observed that SKR has achieved a maximum 2.25\% performance improvement compared to Framework. 
Next, the effectiveness of the proposed DSR was evaluated, which is used to explore the relationships between instance labels and image information under weakly supervised conditions. We observed that this module improved the accuracy by 3.96\% based on ``Framework". 
This suggests that extracting potential similarity relationships in labels will help the model understand features better. 
When SKR and DSR were used together, the model achieved optimal performance.
Since our method uses a noise prediction network, we further validated the model’s efficacy by subjecting it solely to noise addition. We conducted experiments on the CUB dataset and reported the corresponding results. The accuracy for 5-way-1-shot and 5-way-5-shot was 81.72\% and 91.98\%, respectively. Compared to the Framework, there was a decrease of 2.46\% and 1.46\% in accuracy, respectively. This indicates that merely adding noise not only does not help improve the model's generalization ability but also interferes with the model's understanding of fine-grained targets.

\begin{table*}[hbtp]
\caption{Performance combined with the current state-of-the-art FSL methods.}
  \centering  
  \fontsize{1.5}{2}\small
  \begin{threeparttable}  
 \begin{tabular}{c|cc|cc|cc}  
\toprule  
 \multirow{2}{*}{\textbf{Method}}  &\multicolumn{2}{c|}{\bf CUB-200-2011}   &\multicolumn{2}{c|}{\bf Stanford Dogs} &\multicolumn{2}{c}{\bf Stanford Cars}\\
    &\textbf{1-shot}   & \textbf{5-shot}  &\textbf{1-shot}   & \textbf{5-shot}  &\textbf{1-shot}   & \textbf{5-shot}  \\ 
\midrule   

ProtoNet~\cite{snell2017prototypical}~(NIPS@2017) &77.66$\pm$0.21 & 89.42$\pm$0.12 &45.92$\pm$0.21  &67.50$\pm$0.17  &47.60$\pm$0.21  &72.81$\pm$0.18 \\
ProtoNet + DRDM                                   &82.58$\pm$0.18 & 90.80$\pm$0.11 &58.92$\pm$0.18  &70.45$\pm$0.16  &61.66$\pm$0.20  &74.37$\pm$0.17 \\

FRN~\cite{wertheimer2021few}~(CVPR@2021) &83.55$\pm$0.19  &92.92$\pm$0.10  &55.49$\pm$0.21  &74.54$\pm$0.16  &62.07$\pm$0.22  &83.18$\pm$0.14 \\
FRN + DRDM                               &85.40$\pm$0.17  &93.52$\pm$0.10  &61.88$\pm$0.22  &77.86$\pm$0.15  &79.25$\pm$0.19  &88.23$\pm$0.10 \\

\bottomrule
    \end{tabular}  
    \end{threeparttable}  
 \label{boostforothers}
\vspace{-0.2cm} 
\end{table*}

\subsection{Scalability Analysis}
We also apply the augmented data from this paper to existing models to evaluate the gains of our approach over existing algorithms. It can be observed in Table~\ref{boostforothers}, on the CUB dataset, our algorithm achieves a top-1 performance improvement of 4.92\% and 1.85\% for ProtoNet and FRN, respectively. Similar conclusions can be drawn for the Dogs and Cars datasets. This indicates that our algorithm has good portability, which can effectively enhance the performance of other few-shot learning models.

\begin{figure}[hbpt]
  \begin{center}
  \includegraphics[width=0.48\textwidth]{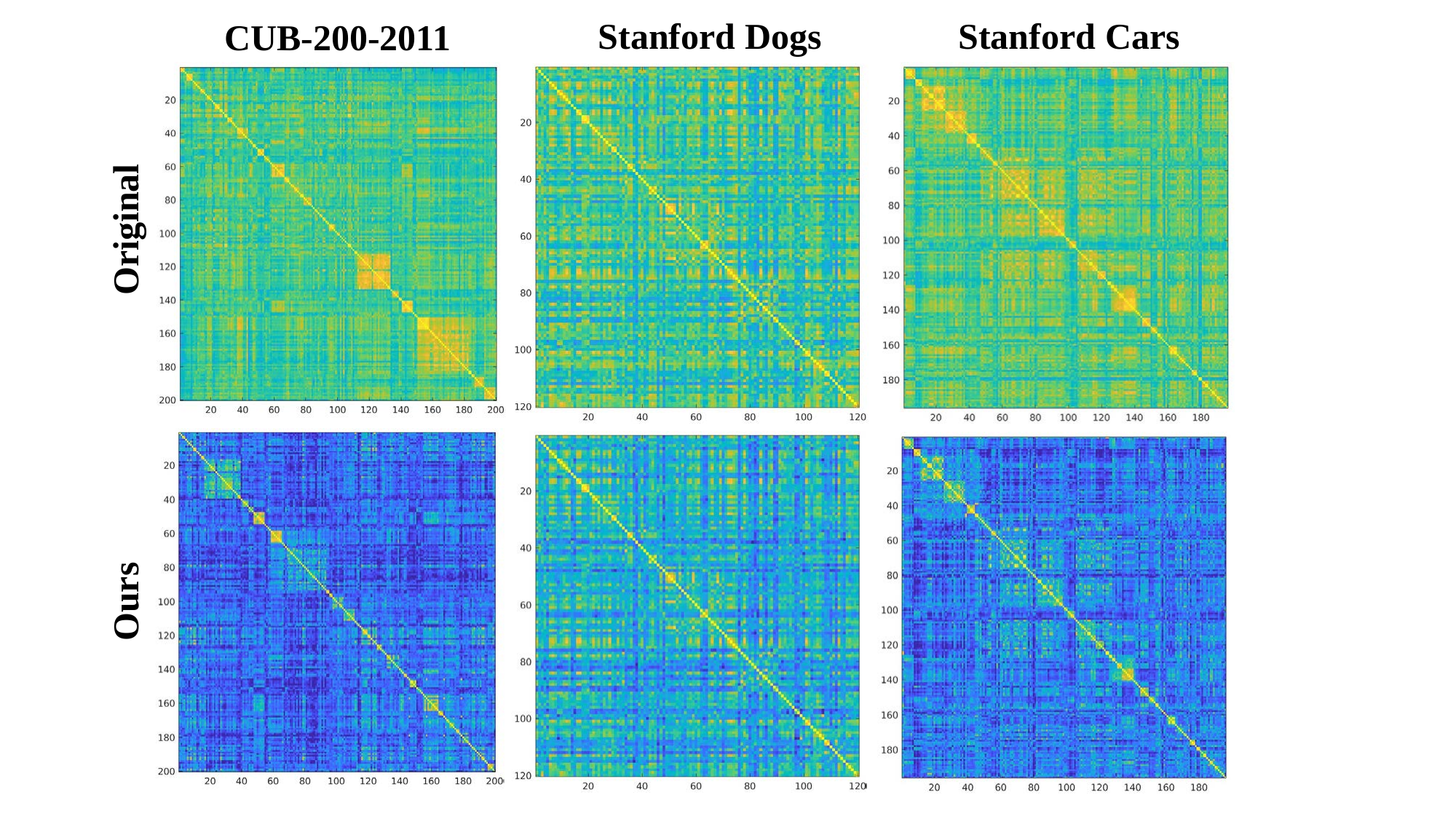}
  \end{center}
	\setlength{\abovecaptionskip}{-0.1cm} 
   \caption{The first and second rows represent the results before and after using DSR, respectively. It can be observed that the effective separation of textual relationships has been achieved.}
\vspace{-0.1cm} 
  \label{similarity_of_label}
\end{figure}

Simultaneously, we visualize the similarity of label features on three datasets to illustrate the improvement of our DSR model in addressing semantic misalignment issues. The results are presented in Figure~\ref{similarity_of_label}.
It can be observed that after applying DSR, the similarity relationships between labels are effectively constrained. This also reveals the reasons why augmented data can effectively prevent the generation of irrelevant images.
Similar results can be observed in the other two datasets. This indicates that our method can better guide the learning process to improve the generalization ability of the model. 

Besides, we also assessed the impact of different types of data on the knowledge referencing module. In this process, the 1-shot results were reported in Table~\ref{kno.ref_in_other}.
\renewcommand{\arraystretch}{1.0}
\begin{table}[hbtp]\normalsize
    \centering
    \caption{Impact of data types on knowledge referencing. ``Pet" represents the ``Oxford-IIIT Pet" dataset.}
    \begin{tabular}{l|c|c|c|c}
        \hline
\textbf{Dataset} & \textbf{Baseline}  & \textbf{NABirds}  & \textbf{Pet}  & \textbf{CompCars}   \\  
        \hline
CUB~\cite{wah2011caltech}   &84.18   &\bf 86.43   &  85.34   &  85.09   \\  
Dogs~\cite{khosla2011novel} &61.44   &61.55       &\bf 62.53 &  60.89   \\  
Cars~\cite{krause20133d}	&67.10   &   68.95   & 68.84  &   \bf 71.79 \\
\hline
    \end{tabular}
    \label{kno.ref_in_other}
\end{table}

It was observed that leveraging external data to enhance the model's understanding of instances under limited samples was effective. When instances exhibit substantial dissimilarities, embedding spatial relationships did not significantly impact the model's performance. However, in cases of similar data types, the model demonstrated greater confidence in discrimination.
We also employed t-SNE~\cite{van2008visualizing} to qualitatively illustrate the impact of knowledge referencing on features. Additionally, we utilized spatial utilization ratio $\rho$~\cite{tian2019sosnet} to quantitatively characterize the degree of subclass aggregation. 
$\rho$ represents the ratio of average inter-class distance to average intra-class distance of features. A higher value indicates a more compact feature distribution and a wider decision boundary allows the model to make decisions more confidently.
The results are presented in Figure~\ref{TSNE}.
Compared to the results of the framework, the utilization of label similarity relationships to constrain model learning has led to feature distribution aggregation. In fact, on the Cars and Dogs datasets, the $\rho$ has even doubled. Subsequently, with the introduction of appropriate knowledge as referencing, the feature distribution exhibited further intra-subclass aggregation and wider inter-subclass boundaries, ultimately benefiting the model's final decision-making process. Compared to the ProtoNet and FRN methods, our algorithm still exhibited the highest spatial utilization rate, which partially explains why we achieved the best results.

\section{Conclusions}
Augmenting data with subtle and consistent discriminative features is one effective approach to achieving reliable few-shot FGVC.
In this paper, we introduce a diffusion-based model solution that explores latent instance similarity relationships within labels and employs external data for feature referencing, achieving enhanced data augmentation through detailed reinforcement.
The experimental results validate that solely employing a diffusion-based class generation model for data augmentation cannot effectively address the challenge of few-shot learning. In contrast, our method leverages the constraints of similar relationships within labels and the reference from external knowledge, effectively alleviating label feature contamination and achieving a substantial performance improvement.
We also qualitatively and quantitatively validated the impact of utilizing reference knowledge from different classes on the model. Our approach can effectively extract knowledge from similar subclass data while avoiding the influence of irrelevant subclasses.
Besides, one thing should be noted, our method primarily relies on label similarity for feature constraints, potentially overlooking underlying cross-hierarchical label relationships.  In the future, we aim to explore additional correlated information, such as cross-hierarchical label relationships or structural information in a single-instance feature, to further enhance the model's generalization capabilities.

\ifCLASSOPTIONcompsoc
  \section*{Acknowledgments}
\else
\fi

\ifCLASSOPTIONcaptionsoff
  \newpage
\fi

\bibliographystyle{IEEEtran} 
\bibliography{ref.bib}

\begin{thebibliography}{10}
\providecommand{\url}[1]{#1}
\csname url@samestyle\endcsname
\providecommand{\newblock}{\relax}
\providecommand{\bibinfo}[2]{#2}
\providecommand{\BIBentrySTDinterwordspacing}{\spaceskip=0pt\relax}
\providecommand{\BIBentryALTinterwordstretchfactor}{4}
\providecommand{\BIBentryALTinterwordspacing}{\spaceskip=\fontdimen2\font plus
\BIBentryALTinterwordstretchfactor\fontdimen3\font minus
  \fontdimen4\font\relax}
\providecommand{\BIBforeignlanguage}[2]{{%
\expandafter\ifx\csname l@#1\endcsname\relax
\typeout{** WARNING: IEEEtran.bst: No hyphenation pattern has been}%
\typeout{** loaded for the language `#1'. Using the pattern for}%
\typeout{** the default language instead.}%
\else
\language=\csname l@#1\endcsname
\fi
#2}}
\providecommand{\BIBdecl}{\relax}
\BIBdecl

\bibitem{wah2011caltech}
C.~Wah, S.~Branson, P.~Welinder, P.~Perona, and S.~Belongie, ``The caltech-ucsd
  birds-200-2011 dataset.''\hskip 1em plus 0.5em minus 0.4em\relax California
  Institute of Technology, 2011.

\bibitem{8591961}
J.~Mańdziuk, ``New shades of the vehicle routing problem: Emerging problem
  formulations and computational intelligence solution methods,'' \emph{IEEE
  Transactions on Emerging Topics in Computational Intelligence}, vol.~3,
  no.~3, pp. 230--244, 2019.

\bibitem{sadeghi2020system}
K.~Sadeghi, A.~Banerjee, and S.~K. Gupta, ``A system-driven taxonomy of attacks
  and defenses in adversarial machine learning,'' \emph{IEEE Transactions on
  Emerging Topics in Computational Intelligence}, vol.~4, no.~4, pp. 450--467,
  2020.

\bibitem{yi2022pharmaceutical}
J.~Yi, H.~Zhang, J.~Mao, Y.~Chen, H.~Zhong, and Y.~Wang, ``Pharmaceutical
  foreign particle detection: An efficient method based on adaptive convolution
  and multiscale attention,'' \emph{IEEE Transactions on Emerging Topics in
  Computational Intelligence}, vol.~6, no.~6, pp. 1302--1313, 2022.

\bibitem{9870199}
J.~Du, K.~Guan, Y.~Zhou, Y.~Li, and T.~Wang, ``Parameter-free similarity-aware
  attention module for medical image classification and segmentation,''
  \emph{IEEE Transactions on Emerging Topics in Computational Intelligence},
  vol.~7, no.~3, pp. 845--857, 2023.

\bibitem{ye2023image}
S.~Ye, Y.~Wang, Q.~Peng, X.~You, and C.~P. Chen, ``The image data and backbone
  in weakly supervised fine-grained visual categorization: A revisit and
  further thinking,'' \emph{IEEE Transactions on Circuits and Systems for Video
  Technology}, 2023.

\bibitem{he2018stackdrl}
X.~He, Y.~Peng, and J.~Zhao, ``Stackdrl: Stacked deep reinforcement learning
  for fine-grained visual categorization.'' in \emph{IJCAI}, 2018, pp.
  741--747.

\bibitem{zheng2020fine}
X.~Zheng, L.~Qi, Y.~Ren, and X.~Lu, ``Fine-grained visual categorization by
  localizing object parts with single image,'' \emph{IEEE Transactions on
  Multimedia}, vol.~23, pp. 1187--1199, 2020.

\bibitem{ding2019selective}
Y.~Ding, Y.~Zhou, Y.~Zhu, Q.~Ye, and J.~Jiao, ``Selective sparse sampling for
  fine-grained image recognition,'' in \emph{Proceedings of the IEEE/CVF
  International Conference on Computer Vision}, 2019, pp. 6599--6608.

\bibitem{ye2022discriminative}
S.~Ye, Q.~Peng, W.~Sun, J.~Xu, Y.~Wang, X.~You, and Y.-M. Cheung,
  ``Discriminative suprasphere embedding for fine-grained visual
  categorization,'' \emph{IEEE Transactions on Neural Networks and Learning
  Systems}, 2022.

\bibitem{wang2022r2}
Y.~Wang, S.~Ye, S.~Yu, and X.~You, ``R2-trans: Fine-grained visual
  categorization with redundancy reduction,'' \emph{arXiv preprint
  arXiv:2204.10095}, 2022.

\bibitem{hong2022semantic}
Z.~Hong, S.~Chen, G.~Xie, W.~Yang, J.~Zhao, Y.~Shao, Q.~Peng, and X.~You,
  ``Semantic compression embedding for generative zero-shot learning,''
  \emph{IJCAI, Vienna, Austria}, vol.~7, pp. 956--963, 2022.

\bibitem{shu2022improving}
Y.~Shu, B.~Yu, H.~Xu, and L.~Liu, ``Improving fine-grained visual recognition
  in low data regimes via self-boosting attention mechanism,'' in
  \emph{Computer Vision--ECCV 2022: 17th European Conference, Tel Aviv, Israel,
  October 23--27, 2022, Proceedings, Part XXV}.\hskip 1em plus 0.5em minus
  0.4em\relax Springer, 2022, pp. 449--465.

\bibitem{10234694}
Y.~Wang, X.~Pei, and H.~Zhan, ``Fine-grained graph learning for multi-view
  subspace clustering,'' \emph{IEEE Transactions on Emerging Topics in
  Computational Intelligence}, pp. 1--12, 2023.

\bibitem{10115455}
K.-Y. Feng, M.~Gong, K.~Pan, H.~Zhao, Y.~Wu, and K.~Sheng, ``Model
  sparsification for communication-efficient multi-party learning via
  contrastive distillation in image classification,'' \emph{IEEE Transactions
  on Emerging Topics in Computational Intelligence}, pp. 1--14, 2023.

\bibitem{li2020adversarial}
K.~Li, Y.~Zhang, K.~Li, and Y.~Fu, ``Adversarial feature hallucination networks
  for few-shot learning,'' in \emph{Proceedings of the IEEE/CVF Conference on
  Computer Vision and Pattern Recognition}, 2020, pp. 13\,470--13\,479.

\bibitem{le2019shadow}
H.~Le and D.~Samaras, ``Shadow removal via shadow image decomposition,'' in
  \emph{Proceedings of the IEEE/CVF International Conference on Computer
  Vision}, 2019, pp. 8578--8587.

\bibitem{zhang2018metagan}
R.~Zhang, T.~Che, Z.~Ghahramani, Y.~Bengio, and Y.~Song, ``Metagan: An
  adversarial approach to few-shot learning,'' \emph{Advances in Neural
  Information Processing Systems}, vol.~31, 2018.

\bibitem{gao2018low}
H.~Gao, Z.~Shou, A.~Zareian, H.~Zhang, and S.-F. Chang, ``Low-shot learning via
  covariance-preserving adversarial augmentation networks,'' \emph{Advances in
  Neural Information Processing Systems}, vol.~31, 2018.

\bibitem{tsutsui2019meta}
S.~Tsutsui, Y.~Fu, and D.~Crandall, ``Meta-reinforced synthetic data for
  one-shot fine-grained visual recognition,'' \emph{Advances in Neural
  Information Processing Systems}, vol.~32, 2019.

\bibitem{rombach2022high}
R.~Rombach, A.~Blattmann, D.~Lorenz, P.~Esser, and B.~Ommer, ``High-resolution
  image synthesis with latent diffusion models,'' in \emph{Proceedings of the
  IEEE/CVF conference on computer vision and pattern recognition}, 2022, pp.
  10\,684--10\,695.

\bibitem{li2023libfewshot}
W.~Li, Z.~Wang, X.~Yang, C.~Dong, P.~Tian, T.~Qin, J.~Huo, Y.~Shi, L.~Wang,
  Y.~Gao \emph{et~al.}, ``Libfewshot: A comprehensive library for few-shot
  learning,'' \emph{IEEE Transactions on Pattern Analysis and Machine
  Intelligence}, 2023.

\bibitem{yan2023clip}
S.~Yan, N.~Dong, L.~Zhang, and J.~Tang, ``Clip-driven fine-grained text-image
  person re-identification,'' \emph{IEEE Transactions on Image Processing},
  2023.

\bibitem{xin2024few}
Z.~Xin, S.~Chen, T.~Wu, Y.~Shao, W.~Ding, and X.~You, ``Few-shot object
  detection: Research advances and challenges,'' \emph{Information Fusion}, p.
  102307, 2024.

\bibitem{snell2017prototypical}
J.~Snell, K.~Swersky, and R.~Zemel, ``Prototypical networks for few-shot
  learning,'' \emph{Advances in Neural Information Processing Systems},
  vol.~30, 2017.

\bibitem{sung2018learning}
F.~Sung, Y.~Yang, L.~Zhang, T.~Xiang, P.~H. Torr, and T.~M. Hospedales,
  ``Learning to compare: Relation network for few-shot learning,'' in
  \emph{Proceedings of the IEEE Conference on Computer Vision and Pattern
  Recognition}, 2018, pp. 1199--1208.

\bibitem{tang2020revisiting}
L.~Tang, D.~Wertheimer, and B.~Hariharan, ``Revisiting pose-normalization for
  fine-grained few-shot recognition,'' in \emph{Proceedings of the IEEE/CVF
  Conference on Computer Vision and Pattern Recognition}, 2020, pp.
  14\,352--14\,361.

\bibitem{chen2019closer}
W.-Y. Chen, Y.-C. Liu, Z.~Kira, Y.-C.~F. Wang, and J.-B. Huang, ``A closer look
  at few-shot classification,'' \emph{arXiv preprint arXiv:1904.04232}, 2019.

\bibitem{tian2020rethinking}
Y.~Tian, Y.~Wang, D.~Krishnan, J.~B. Tenenbaum, and P.~Isola, ``Rethinking
  few-shot image classification: a good embedding is all you need?'' in
  \emph{Computer Vision--ECCV 2020: 16th European Conference, Glasgow, UK,
  August 23--28, 2020, Proceedings, Part XIV 16}.\hskip 1em plus 0.5em minus
  0.4em\relax Springer, 2020, pp. 266--282.

\bibitem{dhillon2019baseline}
G.~S. Dhillon, P.~Chaudhari, A.~Ravichandran, and S.~Soatto, ``A baseline for
  few-shot image classification,'' \emph{arXiv preprint arXiv:1909.02729},
  2019.

\bibitem{yin2018feature}
X.~Yin, X.~Yu, K.~Sohn, X.~Liu, and M.~Chandraker, ``Feature transfer learning
  for deep face recognition with under-represented data,'' \emph{arXiv preprint
  arXiv:1803.09014}, 2018.

\bibitem{hariharan2017low}
B.~Hariharan and R.~Girshick, ``Low-shot visual recognition by shrinking and
  hallucinating features,'' in \emph{Proceedings of the IEEE International
  Conference on Computer Vision}, 2017, pp. 3018--3027.

\bibitem{schwartz2018delta}
E.~Schwartz, L.~Karlinsky, J.~Shtok, S.~Harary, M.~Marder, A.~Kumar, R.~Feris,
  R.~Giryes, and A.~Bronstein, ``Delta-encoder: an effective sample synthesis
  method for few-shot object recognition,'' \emph{Advances in Neural
  Information Processing Systems}, vol.~31, 2018.

\bibitem{xu2021variational}
J.~Xu, H.~Le, M.~Huang, S.~Athar, and D.~Samaras, ``Variational feature
  disentangling for fine-grained few-shot classification,'' in
  \emph{Proceedings of the IEEE/CVF International Conference on Computer
  Vision}, 2021, pp. 8812--8821.

\bibitem{zhang2021prototype}
B.~Zhang, X.~Li, Y.~Ye, Z.~Huang, and L.~Zhang, ``Prototype completion with
  primitive knowledge for few-shot learning,'' in \emph{Proceedings of the
  IEEE/CVF Conference on Computer Vision and Pattern Recognition}, 2021, pp.
  3754--3762.

\bibitem{lee2022task}
S.~Lee, W.~Moon, and J.-P. Heo, ``Task discrepancy maximization for
  fine-grained few-shot classification,'' in \emph{Proceedings of the IEEE/CVF
  Conference on Computer Vision and Pattern Recognition}, 2022, pp. 5331--5340.

\bibitem{ye2023coping}
``Coping with change: Learning invariant and minimum sufficient representations
  for fine-grained visual categorization,'' \emph{Computer Vision and Image
  Understanding}, vol. 237, p. 103837, 2023.

\bibitem{hong2024improving}
\BIBentryALTinterwordspacing
Z.~Hong, Z.~Wang, L.~Shen, Y.~Yao, Z.~Huang, S.~Chen, C.~Yang, M.~Gong, and
  T.~Liu, ``Improving non-transferable representation learning by harnessing
  content and style,'' in \emph{The Twelfth International Conference on
  Learning Representations}, 2024. [Online]. Available:
  \url{https://openreview.net/forum?id=FYKVPOHCpE}
\BIBentrySTDinterwordspacing

\bibitem{ronneberger2015u}
O.~Ronneberger, P.~Fischer, and T.~Brox, ``U-net: Convolutional networks for
  biomedical image segmentation,'' in \emph{Medical Image Computing and
  Computer-Assisted Intervention}.\hskip 1em plus 0.5em minus 0.4em\relax
  Springer, 2015, pp. 234--241.

\bibitem{dosovitskiy2020image}
A.~Dosovitskiy, L.~Beyer, A.~Kolesnikov, D.~Weissenborn, X.~Zhai,
  T.~Unterthiner, M.~Dehghani, M.~Minderer, G.~Heigold, S.~Gelly \emph{et~al.},
  ``An image is worth 16x16 words: Transformers for image recognition at
  scale,'' \emph{arXiv preprint arXiv:2010.11929}, 2020.

\bibitem{liu2019mtfh}
X.~Liu, Z.~Hu, H.~Ling, and Y.-m. Cheung, ``Mtfh: A matrix tri-factorization
  hashing framework for efficient cross-modal retrieval,'' \emph{IEEE
  Transactions on Pattern Analysis and Machine Intelligence}, vol.~43, no.~3,
  pp. 964--981, 2019.

\bibitem{liu2021fddh}
X.~Liu, X.~Wang, and Y.-m. Cheung, ``Fddh: Fast discriminative discrete hashing
  for large-scale cross-modal retrieval,'' \emph{IEEE Transactions on Neural
  Networks and Learning Systems}, vol.~33, no.~11, pp. 6306--6320, 2021.

\bibitem{peng2023towards}
S.-J. Peng, Y.~Fan, Y.-m. Cheung, X.~Liu, Z.~Cui, and T.~Li, ``Towards
  efficient cross-modal anomaly detection using triple-adaptive network and
  bi-quintuple contrastive learning,'' \emph{IEEE Transactions on Emerging
  Topics in Computational Intelligence}, 2023.

\bibitem{baykal2023protodiffusion}
G.~Baykal, H.~F. Karagoz, T.~Binhuraib, and G.~Unal, ``Protodiffusion:
  Classifier-free diffusion guidance with prototype learning,'' \emph{arXiv
  preprint arXiv:2307.01924}, 2023.

\bibitem{pfeiffer2020adapterfusion}
J.~Pfeiffer, A.~Kamath, A.~R{\"u}ckl{\'e}, K.~Cho, and I.~Gurevych,
  ``Adapterfusion: Non-destructive task composition for transfer learning,''
  \emph{arXiv preprint arXiv:2005.00247}, 2020.

\bibitem{ruckle2020adapterdrop}
A.~R{\"u}ckl{\'e}, G.~Geigle, M.~Glockner, T.~Beck, J.~Pfeiffer, N.~Reimers,
  and I.~Gurevych, ``Adapterdrop: On the efficiency of adapters in
  transformers,'' \emph{arXiv preprint arXiv:2010.11918}, 2020.

\bibitem{wang2020k}
R.~Wang, D.~Tang, N.~Duan, Z.~Wei, X.~Huang, G.~Cao, D.~Jiang, M.~Zhou
  \emph{et~al.}, ``K-adapter: Infusing knowledge into pre-trained models with
  adapters,'' \emph{arXiv preprint arXiv:2002.01808}, 2020.

\bibitem{radford2021learning}
A.~Radford, J.~W. Kim, C.~Hallacy, A.~Ramesh, G.~Goh, S.~Agarwal, G.~Sastry,
  A.~Askell, P.~Mishkin, J.~Clark \emph{et~al.}, ``Learning transferable visual
  models from natural language supervision,'' in \emph{International Conference
  on Machine Learning}.\hskip 1em plus 0.5em minus 0.4em\relax PMLR, 2021, pp.
  8748--8763.

\bibitem{wang2019multi}
X.~Wang, X.~Han, W.~Huang, D.~Dong, and M.~R. Scott, ``Multi-similarity loss
  with general pair weighting for deep metric learning,'' in \emph{Proceedings
  of the IEEE/CVF conference on computer vision and pattern recognition}, 2019,
  pp. 5022--5030.

\bibitem{khosla2011novel}
A.~Khosla, N.~Jayadevaprakash, B.~Yao, and F.-F. Li, ``Novel dataset for
  fine-grained image categorization: Stanford dogs,'' in \emph{Proc. CVPR
  workshop on fine-grained visual categorization (FGVC)}, vol.~2, no.~1.\hskip
  1em plus 0.5em minus 0.4em\relax Citeseer, 2011.

\bibitem{krause20133d}
J.~Krause, M.~Stark, J.~Deng, and L.~Fei-Fei, ``3d object representations for
  fine-grained categorization,'' in \emph{Proceedings of the IEEE International
  Conference on Computer Vision Workshops}, 2013, pp. 554--561.

\bibitem{wertheimer2021few}
D.~Wertheimer, L.~Tang, and B.~Hariharan, ``Few-shot classification with
  feature map reconstruction networks,'' in \emph{Proceedings of the IEEE/CVF
  Conference on Computer Vision and Pattern Recognition}, 2021, pp. 8012--8021.

\bibitem{li2019revisiting}
W.~Li, L.~Wang, J.~Xu, J.~Huo, Y.~Gao, and J.~Luo, ``Revisiting local
  descriptor based image-to-class measure for few-shot learning,'' in
  \emph{Proceedings of the IEEE/CVF Conference on Computer Vision and Pattern
  Recognition}, 2019, pp. 7260--7268.

\bibitem{van2015building}
G.~Van~Horn, S.~Branson, R.~Farrell, S.~Haber, J.~Barry, P.~Ipeirotis,
  P.~Perona, and S.~Belongie, ``Building a bird recognition app and large scale
  dataset with citizen scientists: The fine print in fine-grained dataset
  collection,'' in \emph{Proceedings of the IEEE Conference on Computer Vision
  and Pattern Recognition}, 2015, pp. 595--604.

\bibitem{parkhi2012cats}
O.~M. Parkhi, A.~Vedaldi, A.~Zisserman, and C.~Jawahar, ``Cats and dogs,'' in
  \emph{Proceedings of the IEEE Conference on Computer Vision and Pattern
  Recognition}.\hskip 1em plus 0.5em minus 0.4em\relax IEEE, 2012, pp.
  3498--3505.

\bibitem{yang2015large}
L.~Yang, P.~Luo, C.~Change~Loy, and X.~Tang, ``A large-scale car dataset for
  fine-grained categorization and verification,'' in \emph{Proceedings of the
  IEEE conference on computer vision and pattern recognition}, 2015, pp.
  3973--3981.

\bibitem{he2016deep}
K.~He, X.~Zhang, S.~Ren, and J.~Sun, ``Deep residual learning for image
  recognition,'' in \emph{Proceedings of the IEEE Conference on Computer Vision
  and Pattern Recognition}, 2016, pp. 770--778.

\bibitem{sun2019meta}
Q.~Sun, Y.~Liu, T.-S. Chua, and B.~Schiele, ``Meta-transfer learning for
  few-shot learning,'' in \emph{Proceedings of the IEEE/CVF Conference on
  Computer Vision and Pattern Recognition}, 2019, pp. 403--412.

\bibitem{lee2019meta}
K.~Lee, S.~Maji, A.~Ravichandran, and S.~Soatto, ``Meta-learning with
  differentiable convex optimization,'' in \emph{Proceedings of the IEEE/CVF
  conference on computer vision and pattern recognition}, 2019, pp.
  10\,657--10\,665.

\bibitem{mangla2020charting}
P.~Mangla, N.~Kumari, A.~Sinha, M.~Singh, B.~Krishnamurthy, and V.~N.
  Balasubramanian, ``Charting the right manifold: Manifold mixup for few-shot
  learning,'' in \emph{Proceedings of the IEEE/CVF winter conference on
  applications of computer vision}, 2020, pp. 2218--2227.

\bibitem{liu2020negative}
B.~Liu, Y.~Cao, Y.~Lin, Q.~Li, Z.~Zhang, M.~Long, and H.~Hu, ``Negative margin
  matters: Understanding margin in few-shot classification,'' in \emph{Computer
  Vision--ECCV 2020: 16th European Conference, Glasgow, UK, August 23--28,
  2020, Proceedings, Part IV 16}.\hskip 1em plus 0.5em minus 0.4em\relax
  Springer, 2020, pp. 438--455.

\bibitem{afrasiyabi2020associative}
A.~Afrasiyabi, J.-F. Lalonde, and C.~Gagn{\'e}, ``Associative alignment for
  few-shot image classification,'' in \emph{Computer Vision--ECCV 2020: 16th
  European Conference, Glasgow, UK, August 23--28, 2020, Proceedings, Part V
  16}.\hskip 1em plus 0.5em minus 0.4em\relax Springer, 2020, pp. 18--35.

\bibitem{zhu2020multi}
Y.~Zhu, C.~Liu, and S.~Jiang, ``Multi-attention meta learning for few-shot
  fine-grained image recognition.'' in \emph{IJCAI}, 2020, pp. 1090--1096.

\bibitem{li2020bsnet}
X.~Li, J.~Wu, Z.~Sun, Z.~Ma, J.~Cao, and J.-H. Xue, ``Bsnet: Bi-similarity
  network for few-shot fine-grained image classification,'' \emph{IEEE
  Transactions on Image Processing}, vol.~30, pp. 1318--1331, 2020.

\bibitem{tang2022learning}
H.~Tang, C.~Yuan, Z.~Li, and J.~Tang, ``Learning attention-guided pyramidal
  features for few-shot fine-grained recognition,'' \emph{Pattern Recognition},
  vol. 130, p. 108792, 2022.

\bibitem{li2022coarse}
P.~Li, G.~Zhao, and X.~Xu, ``Coarse-to-fine few-shot classification with deep
  metric learning,'' \emph{Information Sciences}, vol. 610, pp. 592--604, 2022.

\bibitem{munjal2023query}
B.~Munjal, A.~Flaborea, S.~Amin, F.~Tombari, and F.~Galasso, ``Query-guided
  networks for few-shot fine-grained classification and person search,''
  \emph{Pattern Recognition}, vol. 133, p. 109049, 2023.

\bibitem{sun2023t2l}
N.~Sun and P.~Yang, ``T2l: Trans-transfer learning for few-shot fine-grained
  visual categorization with extended adaptation,'' \emph{Knowledge-Based
  Systems}, vol. 264, p. 110329, 2023.

\bibitem{pan2023few}
M.-H. Pan, H.-Y. Xin, C.-Q. Xia, and H.-B. Shen, ``Few-shot classification with
  task-adaptive semantic feature learning,'' \emph{Pattern Recognition}, vol.
  141, p. 109594, 2023.

\bibitem{gu2022vector}
S.~Gu, D.~Chen, J.~Bao, F.~Wen, B.~Zhang, D.~Chen, L.~Yuan, and B.~Guo,
  ``Vector quantized diffusion model for text-to-image synthesis,'' in
  \emph{Proceedings of the IEEE/CVF Conference on Computer Vision and Pattern
  Recognition}, 2022, pp. 10\,696--10\,706.

\bibitem{van2008visualizing}
L.~Van~der Maaten and G.~Hinton, ``Visualizing data using t-sne.''
  \emph{Journal of Machine Learning Research}, vol.~9, no.~11, 2008.

\bibitem{tian2019sosnet}
Y.~Tian, X.~Yu, B.~Fan, F.~Wu, H.~Heijnen, and V.~Balntas, ``Sosnet: Second
  order similarity regularization for local descriptor learning,'' in
  \emph{Proceedings of the IEEE/CVF Conference on Computer Vision and Pattern
  Recognition}, 2019, pp. 11\,016--11\,025.

\end{thebibliography}

\vspace{-0.8cm}

\begin{IEEEbiography}
[{\includegraphics[width=1in,height=1.25in,clip,keepaspectratio]{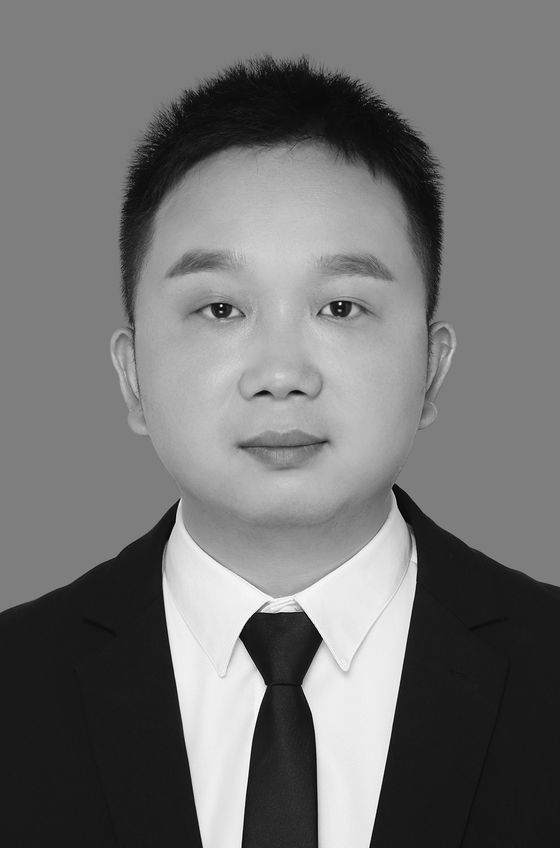}}]{Tianxu Wu}
is currently pursuing the Ph.D. in School of Electronic Information and Communications, Huazhong University of Science and Technology, Wuhan, China. His research interests include machine learning and computer vision.
\end{IEEEbiography}
\vspace{-1.8cm}

\begin{IEEEbiography}
[{\includegraphics[width=1in,height=1.25in,clip,keepaspectratio]{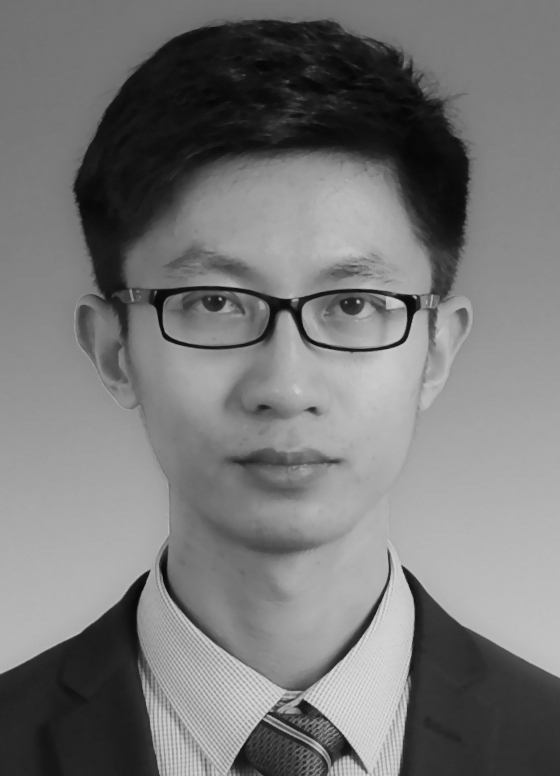}}]{Shuo Ye}
is currently a full-time Ph.D student in the School of Electronic Information and Communications, Huazhong University of Sciences and Technology (HUST), China. His current research interests span computer vision and voice signal processing with a series of topics, such as automatic speech recognition and fine-grained image categorization. 
\end{IEEEbiography}
\vspace{-1.8cm}

\begin{IEEEbiography}[{\includegraphics[width=1in,height=1.25in,clip,keepaspectratio]{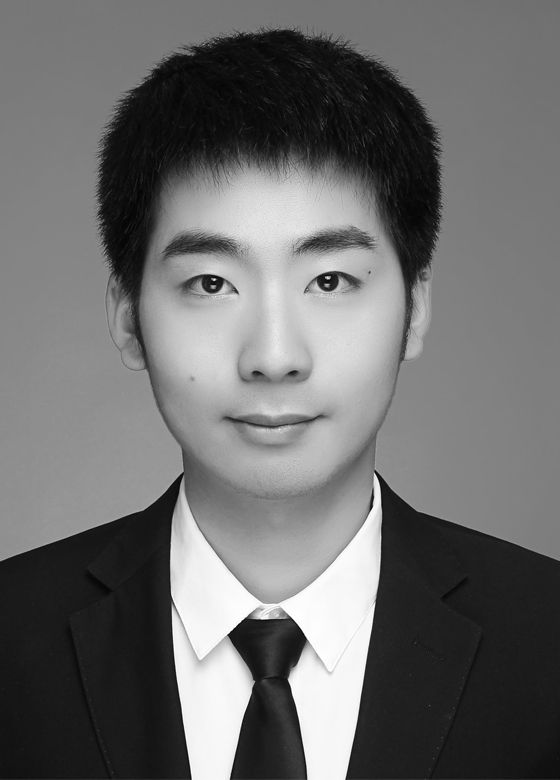}}]{Shuhuang Chen}
is currently pursuing the Ph.D. in School of Electronic Information and Communications, Huazhong University of Science and Technology, Wuhan, China. His research interests include machine learning and computer vision.
\end{IEEEbiography}

\vspace{-1.81cm}

\begin{IEEEbiography}[{\includegraphics[width=1in,height=1.25in,clip,keepaspectratio]{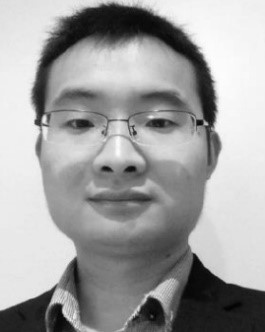}}]{Qinmu Peng}
received the Ph.D. degree from the Department of Computer Science, Hong Kong Baptist University, Hong Kong, in 2015. He is currently an Assistant Professor with the School of Electronic Information and Communications, Huazhong University of Science and Technology, Wuhan, China, and the Shenzhen Research Institute, Huazhong University of Science and Technology, Shenzhen, China. His current research interests include medical image processing, pattern recognition, machine learning, and computer vision.
\end{IEEEbiography}

\vspace{-1.81cm}
\begin{IEEEbiography}[{\includegraphics[width=1in,height=1.25in,clip,keepaspectratio]{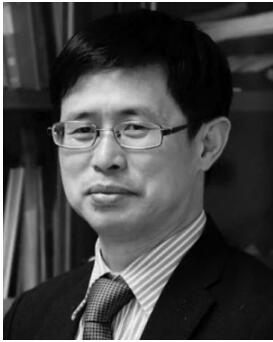}}]{Xinge You}
(M’08-SM’10) received the BS and MS degrees in mathematics from the Hubei University, Wuhan, China and the PhD degree from the Department of Computer Science, Hong Kong Baptist University, Hong Kong, in 1990, 2000, and 2004, respectively. Currently, he is a professor with the School of Electronic Information and Communications, Huazhong University of Science and Technology, China. His research interests include pattern recognition, image and signal processing, computer vision and machine learning. He has published more than 100 papers, such as the IEEE Transactions on Pattern Analysis and Machine Intelligence, TCB, the IEEE Transactions on Image Processing and CVPR. He is a senior member of the IEEE.
\end{IEEEbiography}

\end{document}